\documentclass{article}
\usepackage{ijcai22}

\usepackage{times}
\usepackage{soul}
\usepackage{url}
\usepackage[hidelinks]{hyperref}
\usepackage[utf8]{inputenc}
\usepackage[small]{caption}
\usepackage{graphicx}

\usepackage{amsthm}
\usepackage{booktabs}
\usepackage{algorithm}
\usepackage{algorithmic}
\usepackage{subfigure} 
\usepackage{thmtools} 
\usepackage{thm-restate} 
\usepackage{multirow}
\usepackage{multicol}

\newtheorem{lemma}{Lemma}
\newtheorem{remark}{Remark}
\newtheorem{definition}{Definition}
\newtheorem{proposition}{Proposition}

\usepackage{amsmath,amsfonts}
\usepackage[dvipsnames]{xcolor}

\newcommand{\citet}[1]{\cite{#1}}
\newcommand{\citep}[1]{\cite{#1}}
\newcommand{\appref}[1]{\ref{#1}}
\newcommand{\appcite}[1]{\cite{#1}}

\newcommand{\appcitet}[1]{\cite{#1}}

\newcommand{\mbf}[1]{\mathbf{#1}}

\newcommand{\mcal}[1]{\mathcal{#1}}

\def\W{\mathbf{W}}
\def\L{\mathbf{L}}
\usepackage{amssymb} %
\usepackage{bbm} %
\def\R{\mathbb{R}}

\def\U{\mathbf{U}}
\def\u{\mathbf{u}}

\def\x{\mathbf{x}}
\def\y{\mathbf{y}}
\def\v{\mathbf{v}}
\def\e{\mathbf{e}}

\def\X{\mathbf{X}}

\def\xx{\times}
\def\V{\mathcal{V}}
\def\E{\mathcal{E}}

\def\D{\mbf{D}}
\def\W{\mbf{W}}

\def\L{\mbf{L}}
\def\K{\mbf{K}}
\def\H{\mbf{H}}
\def\P{\mbf{P}}
\def\C{\mbf{C}}
\def\A{\mbf{A}}
\def\a{\mbf{a}}
\def\b{\mbf{b}}
\def\AP{\A\odot\P}
\def\M{\mbf{M}}
\def\MP{\mbf{M}\odot\P}

\def\f{\mathbf{f}}
\def\g{\mathbf{g}}
\def\p{\mathbf{p}}
\def\q{\mathbf{q}}
\usepackage{xspace}  
\newcommand{\mot}{GTOT\xspace}
\usepackage{etoc}
\etocdepthtag.toc{mtchapter}
\etocsettagdepth{mtchapter}{subsection}
\etocsettagdepth{mtappendix}{none}

\usepackage{framed}
\definecolor{link_color}{RGB}{153, 0,0}  
\definecolor{url_color}{RGB}{153, 102,  0}
\definecolor{emp_color}{RGB}{0,0,255}
\definecolor{cite_color}{HTML}{114083}
\hypersetup{
 colorlinks,
 citecolor=cite_color,
 linkcolor=link_color,
 urlcolor=url_color}
\definecolor{shadecolor}{rgb}{0.94, 0.97, 1.0}

\title{
Fine-Tuning Graph Neural Networks via\\
Graph Topology induced Optimal Transport}

\author{
Jiying Zhang$^{1,2}$
\and
Xi Xiao$^2$
\and
Long-Kai Huang$^1$
\and
Yu Rong$^1$
\And
Yatao Bian$^1$\thanks{Correspondence to: Yatao Bian (\texttt{yatao.bian@gmail.com})}
\affiliations
$^1$Tencent AI Lab, Shenzhen, China\\
$^2$Shenzhen International Graduate School, Tsinghua University, Shenzhen, China\\
\emails
zhangjiy20@mails.tsinghua.edu.cn,
xiaox@sz.tsinghua.edu.cn,
yu.rong@hotmail.com,
\{hlongkai,yatao.bian\}@gmail.com
}

\begin{document}

\maketitle

\begin{abstract}



Recently,  the pretrain-finetuning paradigm has attracted tons of attention in graph learning community due to its power of alleviating the lack of labels problem in many real-world applications. 
Current studies use existing techniques, such as weight constraint, representation constraint, which are derived from images or text data, to transfer the invariant knowledge from the pre-train stage to fine-tuning stage.
However, these methods failed to preserve invariances from graph structure and Graph Neural Network (GNN) style models.
In this paper, we present a novel optimal transport-based fine-tuning framework called GTOT-Tuning, namely, Graph Topology induced Optimal Transport fine-Tuning, for GNN style backbones.   
GTOT-Tuning is required to utilize the property of graph data to enhance the preservation of representation produced by fine-tuned networks.
Toward this goal, we formulate graph local knowledge transfer as an Optimal Transport (OT) problem with a structural prior and construct the GTOT regularizer to constrain the fine-tuned model behaviors.
By using the adjacency relationship amongst nodes, the GTOT regularizer achieves node-level optimal transport procedures and reduces redundant transport procedures, resulting in efficient knowledge transfer from the pre-trained models. 
We evaluate GTOT-Tuning on eight downstream tasks with various GNN backbones and demonstrate that it achieves state-of-the-art fine-tuning performance for GNNs.

\end{abstract}

\section{Introduction}

Learning from limited number of training instances is a fundamental problem in many real-word applications. A popular approach to address this issue is fine-tuning a model that pre-trains on a large dataset. In contrast to training from scratch, fine-tuning usually requires fewer labeled data, allows for faster training, and generally achieves better performance~\cite{li2018delta,he2019rethinking}.

Conventional fine-tuning approaches can be roughly divided into two categories: (i) weight constraint~\cite{xuhong2018explicit}, i.e. directly constraining the distance of the weights between pretrained and finetuned models. 
Obviously, they would fail to utilize the topological information of graph data.
(ii) representation constraint~\cite{li2018delta}. This type of approach constrains the distance of representations produced from pretrained and finetuned models, preserving the outputs or intermediate activations of finetuned networks. Therefore, both types of approaches fail to take good account of the topological information implied by the middle layer embeddings. However, it has been proven that GNNs explicitly look into the topological structure of the data by exploring the local and global semantics of the graph~\cite{xu2018powerful,hu2020strategies,xu2021self}, which means that the implicit structure between the node embeddings is very significant.
As a result, these fine-tuning methods, which cover only weights or layer activations and ignore the topological context of the input data, are no longer capable of obtaining comprehensive knowledge transfer.

\begin{figure*}[ht]
    \centering
           \vspace{-3mm}
    \includegraphics[width=0.99\linewidth]{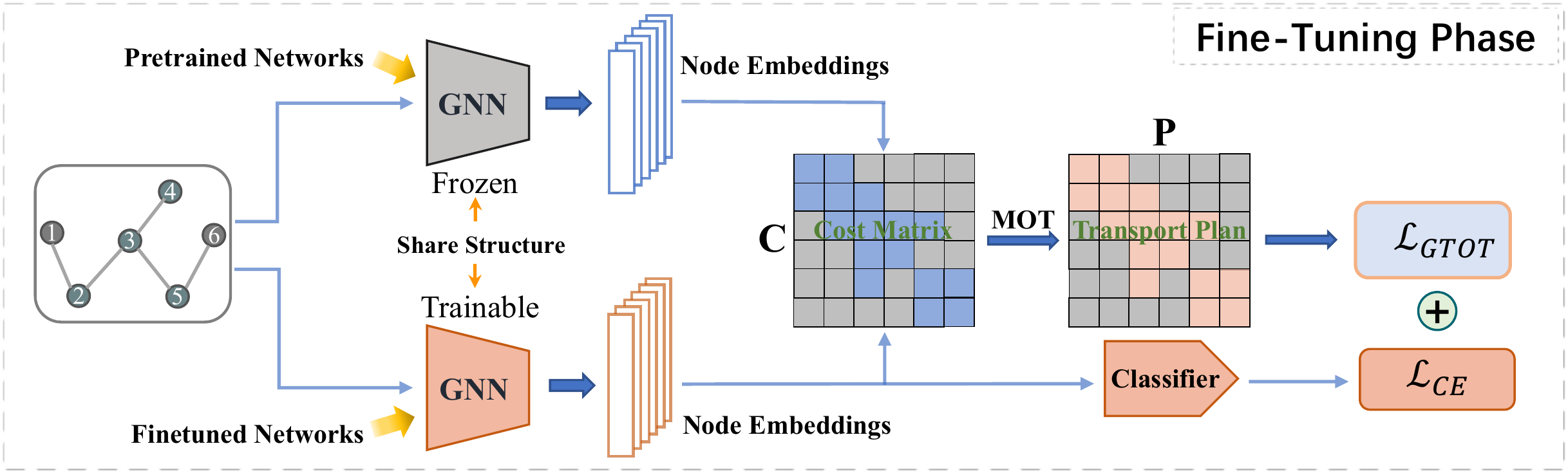}
    \vspace{-2mm}
     \caption{Overall framework of \mot-Tuning, where $\mcal{L}_{GTOT}$ denotes GTOT regularizer loss and $\mcal{L}_{CE}$ represents Cross Entropy loss. The gray lattice of $\P$ indicates that
     $\P_{ij}=0$ when the vertex pair $(v_i,v_j)$ is not adjacent. Assume that the input graph has self-loops.  }
    
    \label{fig:gtot_framework}
        \vspace{-4mm}
\end{figure*}

To preserve the local information of finetuned network from pretrained models, in this paper, we explore a principled representation regularization approach. i) Masked Optimal Transport~(MOT) is formalized and used as a knowledge transfer procedure between pretrained and finetuned models. Compared with the Typical OT distance~\cite{peyre2020computational}, which considers all pair-wise distance between two domains \cite{courty2016optimal}, MOT allows for choosing the specific node pairs to sum in the final OT distance due to the introduced mask matrix.  ii)  The topological information of graphs is incorporated into the Masked Wasserstein distance~(MWD) via setting the mask matrix as an adjacency matrix, leading to a GTOT distance within the node embedding space. The embedding distance between finetuned and pretrained models is minimized by penalizing the GTOT distance, preserving the local information of finetuned models from pretrained models. Finally, we propose a new fine-tuning framework: GTOT-Tuning as  illustrated in Fig. \ref{fig:gtot_framework}.

 Using the adjacency relationship between nodes, the proposed  GTOT regularizer achieves precise node-level optimal transport procedures and omits unnecessary transport procedures, resulting in efficient knowledge transfer from the pretrained models~(Fig. \ref{fig:maskedWD}).
Moreover, thanks to the OT optimization that dynamically updates the transport map~(weights to summing cosine dissimilarity) during training, the proposed regularizer is able to adaptively and implicitly adjust the distance between fine-tuned weights and pre-trained weights according to the downstream task. 
The experiments  conducted on eight different datasets with various GNN backbones show that GTOT-Tuning achieves the best performance among all baselines, validating the effectiveness and generalization ability of our method.

 Our main contributions can be summarized as follows. 

    1) We propose a masked OT problem as an extension of the classical Optimal Transport problem  by introducing a mask matrix to constrain the transport procedures.
    Specially, we define \textit{Masked Wasserstein distance (MWD)} for providing a flexible metric to compare two distributions.
    
    2) We propose a fine-tuning framework called GTOT-Tuning tailored for GNNs, based on the proposed  MWD. The core component of this framework, \textit{GTOT Regularizer}, has the ability to utilize graph structure to preserve the local feature invariances between finetuned and pretrained models.
    To the best of our knowledge, it is the first fine-tuning method tailored for GNNs.
    
    3) Empirically, we conduct extensive experiments on various benchmark datasets, and the results demonstrate a competitive performance of our method.

\section{Related Work}
\paragraph{Pretraining GNNs.}
Pre-training techniques have been shown to be effective for improving the generalization ability of GNN models. The existing methods for pre-training GNNs are mainly based on self-supervised paradigms. Some self-supervised tasks, e.g. context prediction \cite{hu2020strategies,rong2020self}, edge/attribute generation~\cite{hu2020strategies}  and contrastive learning (\cite{you2020graph,xu2021self}), have been designed to 
obtain knowledge from unlabeled graphs.  
However, most of these methods only use the vanilla fine-tuning methods, i.e. the pretrained weights act as the initial weights for downstream tasks. It remains open to exploiting the optimal performance of the pre-trained GNN models. Our work is expected to utilize the graph structure to achieve better performance on the downstream tasks.

\paragraph{Fine-tuning in Transfer learning.}
Fine-tuning a pre-trained model to downstream tasks is a popular paradigm in transfer learning~(TL). \citet{donahue2014decaf,oquab2014learning} indicate that transferring features extracted by pre-trained AlexNet model to downstream tasks yields  better performance  than hand-crafted features.
Further studies by \citet{yosinski2014transferable,agrawal2014analyzing} show that fine-tuning the pre-trained networks is more effective than fixed pre-trained representations.
Recentl research primarily focuses on how to better tap into the prior knowledge of pre-trained models from various perspectives. i) Weights: L2\_SP~\citet{xuhong2018explicit} propose a $L_2$ distance regularization that penalizes $L_2$ distance between the fine-tuned weights and pre-trained weights. 
ii) Features: DELTA \cite{li2018delta} constrains feature maps with the pre-trained activations selected by channel-wise attention. iii) Architecture: 
 BSS~\cite{chen2019catastrophic} penalizes smaller singular values to suppress untransferable spectral components to prevent negative transfer. StochNorm~\citet{kou2020stochastic} uses Stochastic Normalization to replace the classical batch normalization of pre-trained models. Despite the encouraging progress, it still lacks a fine-tuning method specifically for GNNs. 

\paragraph{Optimal Transport.}
Optimal Transport is frequently used in many applications of deep learning, including domain adaption~\cite{courty2016optimal,xu2020reliable}, knowledge distillation~\citep{chen2021wasserstein}, sequence-to-sequence
learning~\citet{chen2019improving}, 
graph matching~\citet{xu2019gromov}, 
cross-domain alignment~\cite{chen2020graph}, 
rigid protein docking \cite{ganea2022independent}, 
and GNN architecture design~\citet{becigneul2020optimal}. Some classical solutions to OT problem like Sinkhorn Algorithm can be found in \citep{peyre2020computational}.
The  work closely related to us may be \cite{li2020representation}, which raises a fine-tuning method based on typical OT. The significant differences are that our approach is i) based on the proposed MOT, ii) tailored for GNNs, and iii) able to exploit the structural information of graphs.

\section{Preliminaries}
\paragraph{Notations.}
We define the inner product $\left<\cdot, \cdot\right>$
for matrices $\mbf{U, V}\in \R^{m\xx n}$ by $\left<\mbf{U},\mbf{V} \right> = \operatorname{tr}(\mbf{U}^{\top}\mbf{V}) = \sum_{i,j} \mbf{U}_{ij}\mbf{V}_{ij} $. $\mbf{I}\in\R^{n\xx n}$ denotes the identity matrix,
and $\mbf 1_n\in \R^{n}$ represents the vector with ones in each component of size $n$. We use boldface letter $\x\in \R^n$
 to indicate an $n$-dimensional vector, where $\x_i$ is
the $i^\text{th}$ entry of $\x$. 
 Let $G(\mcal{V},\mcal{E})$ be a graph with vertices $\V$ and edges $\E$. We denote by $\A\in \R^{|\V|\xx|\V|}$ the adjacency matrix of $G$ and  $\odot$ the Hadamard product.
 
 For convenience, we follow the terms of transfer learning and call the signal graph output from fine-tuned models (target networks) as \textbf{target graph} (with node embeddings $\{x^T_1,... ,x^T_{|\V|}\}$), and correspondingly, the output from pre-trained models~(source networks) is called \textbf{source graph} (with node embeddings $\{x^S_1,... ,x^S_{|\V|}\} $). Note that these two graphs have the same adjacency matrix in fine-tuning setting.
 
\paragraph{Wasserstein Distance.}
Wasserstein distance (WD)~\cite{peyre2020computational} is commonly used for matching two empirical distributions (e.g., two sets of node embeddings in a graph)~\cite{courty2016optimal}. The WD can be defined below.
\begin{definition}
\label{def:WD}
Let $\alpha=\sum_i^n \a_i \delta_{\x_i}$ and $\beta =\sum^m_i \b_i\delta_{\mbf{y}}$ be two discrete distributions with $\delta_{\x_i}$ as the Dirac function concentrated at location $\x$. $\Pi(\alpha,\beta)$  denotes all the joint distributions $\gamma(\x,\y)$, with marginals $\alpha(\x)$ and $\beta(\y)$. $\a\in \R_+^{n}$ and $\b\in \R_+^{m}$ are weight vectors satisfying  $\sum^n_{i=1} \a_i = \sum^m_{i=1} \b_i = 1$. The definition of Wasserstein distance between the two discrete distributions $\alpha$, $\beta$ is:
\begin{align}
    \mathcal{D}_{w}(\alpha,\beta) &= \inf\limits_{\gamma\in\Pi(\alpha,\beta)}\mathbb{E}_{(\x,\y)\sim \gamma}c(\x,\y)
\vspace{-3mm}
\end{align}
or
\vspace{-2mm}
\begin{align}
\small
     \L_{w}(\a,\b)&=\min\limits_{\P\in \mbf{U}(\a,\b)}\left< \P,\C\right>=\min\limits_{\P\in \mbf{U}(\a,\b)}\sum_{ij}\P_{ij}\C_{ij}
\end{align}
where $\mbf{U}(\a,\b) = \{\P \in \R^{n\xx m} \mid \P\mbf{1}_m = \a, \P^{\top}\mbf{1}_n = \b \}$ and $\C_{ij}=c(\x_i,\y_j)$ is a cost scalar representing the distance between $\x_i$ and $\y_j$.
The $\P\in \R^{n\xx m}$ is called as \textbf{{transport plan}}
or \textbf{{transport map}}, and $\P_{ij}$ represents the amount of mass to be moved from $\a_i$ to $\b_j$. $\a,\b$ are also known as marginal distributions of $\P$.

\end{definition}

The discrepancy between each pair of samples across the two domains can be measured by the optimal transport distance $\mathcal{D}_{w}(\alpha,\beta)$. This seems to imply that $\L_{w}(\a,\b)$ is a natural choice as a representation distance between source graph and target graph for GNN fine-tuning. However, the local dependence exists between  source and target graph, especially when the graph is large~(see section \ref{sec:MOT-tuning}). Therefore, it is not appropriate for WD to sum the distances of all node pairs. Inspired by this observation, we propose a masked optimal transport problem, as an extension of typical OT~(section \ref{sec:MOT}).

\begin{algorithm}[htb]
\def\u{\mathbf{u}}
\def\v{\mathbf{v}}
\small
   \caption{Computing Masked Wasserstein Distance}
   \label{alg:gtot}
\begin{algorithmic}
   \STATE {\bfseries Input:} Cost Matrix $\C$, Mask matrix $\M\in \{0,1\}^{n\xx m}$, Marginals  $\a\in\R_+^{n},\b\in\R_+^{m}$, Threshold $\tau$. \\
    \STATE \textbf{Initialize}: $\mathbf{u}=\mathbf{v}=\mathbf{0}$.

   \FOR{$i=1,2,3,...$}
  \STATE \,\quad $\mathbf{u}_1=\u$
   \vspace{-7mm}
   \STATE
\begin{align*}
\vspace{-1 mm}
\raggedleft  \u=  \epsilon(\log(\a)- & \quad\quad  \\
\quad  \log(\M\odot & \exp(-\C+\u\mbf{1}_m^{\top}+\mbf{1}_n\v^{\top})\mbf{1}_m)) + \u  \\
\v= \epsilon (\log(\b)- &\\
~ \log((\M\odot & \exp(-\C+\u\mbf{1}_m^{\top}+\mbf{1}_n\v^{\top}))^{\top}\mbf{1}_n)) + \v 
\end{align*}
    \vspace{-6mm}
   \IF{$\|\mathbf{u}_1-\u \|_1<\tau$}
   \STATE Break
   \ENDIF
   \ENDFOR
   \STATE $\P = \M\odot \exp(-\M\odot \C + \u\mbf{1}_m^{\top} + \mbf{1}_n\v^{\top})$
   \STATE  $\mathcal D_{mw} = \left<\P,\C \right>$
   \STATE {\bfseries Output:} $\P, \mathcal D_{mw}$
   
\end{algorithmic}
\end{algorithm}
\vspace{-2mm}
\section{Masked Optimal Transport}\label{sec:MOT}
Recall that in typical OT (definition \ref{def:WD}), $\a_i$ can be transported to any $\b_j\in \{\b_k\}_{k=1}^{m}$ with amount $\P_{ij}$. Here, we assume that the $\a_i$ can only be transported to  $\b_j \in \mcal{U}$ where $\mcal{U}\subseteq \{\b_k\}_{k=1}^{m}$ is a subset. This constraint can be implemented by limiting the transport plan: $\P_{ij}=0$ if $\b_j\notin \mcal{U}$. Furthermore, the problem is formalized as follows by introducing the mask matrix.
 
\begin{restatable}[Masked Wasserstein distance]{definition}{restadefMWD}
 \label{def:MWD} 
 Following the same notation of definition $\ref{def:WD}$ and given a mask matrix $\M$\footnote{In this paper, $\M_{ij}=0$ represents the $ij$-th element of the matrix being masked.} $\in\{0,1\}^{n\times m}$ where every row or column is not all zeros, the masked Wasserstein distance~(MWD) is defined as
 \begin{align}
     \mathbf{L}_{mw}(\M, \a,\b)= \min\limits_{\P\in \mbf{U}(\M,\a,\b)}\left< \MP,\C\right>, \label{eq:MWD}
\end{align}
where $\mbf{U}(\M,\a,\b):=\{ \P\in\R_+^{n\times m} \mid (\MP) \mbf{1}_m = \a, (\MP)^{\top}\mbf{1}_n = \b, \P\odot(\mbf{1}_{n\xx m}-\M)=\mbf{0}_{n\xx m}\}$ and $\C\in\R^{n\times m}$ is a cost matrix.
 \end{restatable}
From Eq.~\eqref{eq:MWD}, the mask matrix $\M$ indicates that the elements of $\P$ need to be optimized, in other words, the costs need to be involved in the summation when calculating the inner product.  Notably, different mask matrices $\M$ lead to different transport maps and obtain the OT distance associated with $\M$. One can design $\M$ carefully to get a specific WD. Moreover, the defined MWD can recover WD by setting $\M=\mbf{1}_{n\times m}$ and it is obvious that  $\mbf{L}_{mw}(\M,\a,\b)\geq  \mbf{L}_w(\a,\b)$. This problem can obtain approximate solutions by adding an entropic regularization penalty, which is essential for deriving algorithms suitable for parallel iterations~\cite{peyre2020computational}.
\begin{restatable}{proposition}{restadeproMWD}
\label{pro:MWD}
The solution to definition \ref{def:MWD} with entropic regularization $\epsilon\H(\MP)$\footnote{Namely,$\min\limits_{\P\in \mbf{U}(\M,\a,\b)}\left< \MP,\C\right>-\epsilon\H(\MP)$, where $\H(\cdot)$ is Entropy function. Assume that $0\log0=0$ to ensure that $\H(\MP)$ is well defined.} is unique and has the form 
\begin{align}
\label{eq:solver}
    \P_{ij} = \u_i\M_{ij}\K_{ij}\v_j
\end{align}
where $\K_{ij}=\exp(-\C_{ij}/\epsilon)$ and $(\u,\v)\in \R^{n}_+\times \R^{m}_+$ are two unknown scaling variables.
\end{restatable}
It is clear from the result that MWD is not equivalent to weighting the distance matrix $\C$ directly~\cite{xu2020reliable}, so the masked OT problem is nontrivial. We provide the proof in Appendix and the key to it is the observation that $\exp(\M\odot\X) = \M\odot \exp(\X)+{\mbf{1}_{n\xx m}-\M}$, where $\X\in\R^{n\xx m}$ is an arbitrary given matrix. 

A conceptually simple method to compute the solution would be through the Sinkhorn Knopp iterations(Appendix \ref{subsec:MSI}). However, as we know, the Sinkhorn algorithm suffers from numerical overflow when the regularization parameter $\epsilon$ is too small compared to the entries of the cost
matrix $\C$~\cite{peyre2020computational}. This problem may be more severe when the sparsity of the mask matrix is large. Fortunately, this concern can be somewhat alleviated by performing computations in the $\log$ domain. Therefore, for numerical stability and speed, we propose the Log-domain masked Sinkhorn algorithm (the derivation can be  seen in Appendix~\appref{subsec:LD_Masked_Sinkhorn}).
Algorithm \ref{alg:gtot} provides the whole process.

A further extension of the masking idea involves adding a mask matrix to the Gromov-Wasserstein distance~\cite{peyre2016gromov}(MGWD), which can be used to compute distances between pairs of nodes in each domain, as well as to determine how these distances compare with those in the counterpart domain. The definition and the algorithm can be found in Appendix \ref{subsec:MGWD_definition}, \ref{sec:MGWD_regularizer}.

\begin{figure}[t]
    \centering
    \vspace{-1mm}
    \includegraphics[width=0.95\linewidth]{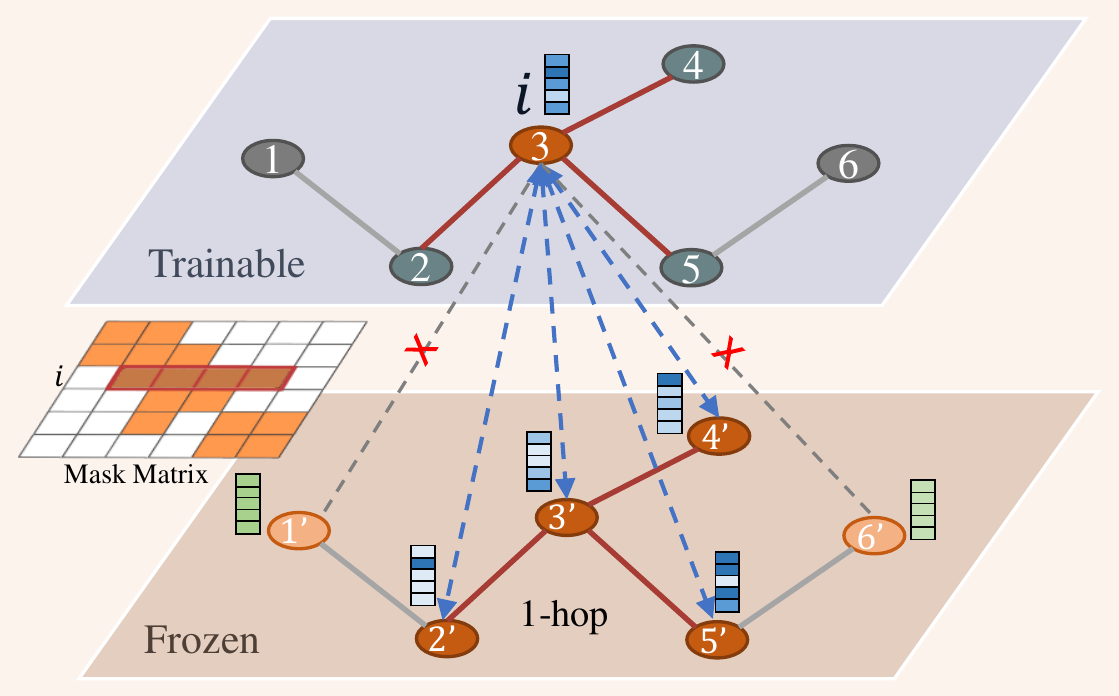}
     \caption{ An example of calculating GTOT distance. When preserving the $i$-th node representation in target graph, the corresponding vertices in the source graph within 1-hop vertices distance  ~(i.e.$\{2',3',4',5'\}$) would be considered. This implies that GTOT regularizer is a local knowledge transfer  regularizer.}
    \label{fig:maskedWD}
        \vspace{-4mm}
\end{figure}

\section{Fine-Tuning GNNs via Masked OT Distance} \label{sec:MOT-tuning}
\label{sec:GTOT}

In our proposed framework, we use Masked Optimal Transport (MOT) for GNNs fine-tuning, where a transport plan $\P \in \R^{n\times m}$ is learned to optimize the knowledge transfer between pretrained and fine-tuned models. There are several distinctive characteristics of MOT that make it an ideal tool for fine-tuning GNNs.
(i) \textbf{Self normalization}: All the elements of $\P$ sum to $1$.
(ii) \textbf{Sparsity}: 
The introduced mask matrix can effectively limit the sparsity of the transport map, leading to a more interpretable and robust representation regularizer for fine-tuning~(Fig. \appref{fig:mwd_transport_map} in Appendix.). 
(iii) \textbf{Efficiency}: Our solution can be easily obtained by Algorithm \ref{alg:gtot} that only requires matrix-vector products, therefore is readily applicable to GNNs.
(iv) \textbf{Flexibility}:
The Masked matrix can assign exclusive transport plans for specific transport tasks and reduce the unnecessary optimal transport procedure.

\begin{table*}[th]
\def\p{$\pm$} 
\setlength\tabcolsep{4pt} 
\centering
\vspace{-3mm}
\caption{Test ROC-AUC (\%) of GIN(contexpred) on downstream molecular property prediction benchmarks.('$\uparrow$' denotes  performance improvement compared to the Fine-Tuning baseline. )}
\vspace{-3mm}
\scalebox{0.74}{
\begin{tabular}{l|llllllll|c}
\toprule 
Methods & BBBP & Tox21 & Toxcast& SIDER & ClinTox & MUV & HIV & BACE & \textbf{Average}\\
\midrule
Fine-Tuning~(baseline) & 68.0$\pm$2.0 & \underline{75.7\p 0.7} & 63.9\p 0.6 & 60.9\p 0.6 & 65.9\p 3.8 & 75.8\p 1.7 & 77.3\p 1.0 & 79.6\p 1.2 & 70.85 \\
\midrule
L2\_SP~\cite{xuhong2018explicit} & 68.2$\pm$0.7 & 73.6$\pm$0.8 & 62.4$\pm$0.3 & 61.1$\pm$0.7 & 68.1$\pm$3.7 & 76.7$\pm$0.9 & {75.7$\pm$1.5} & 82.2$\pm$2.4 & 70.25 \\
DELTA\cite{li2018delta} & 67.8$\pm$0.8 & 75.2$\pm$0.5  & 63.3$\pm$0.5 & 62.2$\pm$0.4 & \textbf{73.4$\pm$3.0} & \textbf{80.2$\pm$1.1} & 77.5$\pm$0.9 & 81.8$\pm$1.1 & 72.68 \\
Feature(DELTA w/o ATT) & 61.4$\pm$0.8 & 71.1$\pm$0.1 & 61.5$\pm$0.2 & \underline{62.4$\pm$0.3} & 64.0$\pm$3.4 & {78.4$\pm$1.1} & 74.0$\pm$0.5 & 76.3$\pm$1.1 & 68.64 \\
BSS\cite{chen2019catastrophic} & 68.1$\pm$1.4 & \textbf{75.9$\pm$0.8} &  \underline{63.9$\pm$0.4} & 60.9$\pm$0.8 & {70.9$\pm$5.1} & 78.0$\pm$2.0 & \underline{77.6$\pm$0.8} & \underline{82.4$\pm$1.8} & 72.21 \\
StochNorm~\cite{kou2020stochastic} & \underline{69.3$\pm$1.6} & 74.9$\pm$0.6 & 63.4$\pm$0.5 & 61.0$\pm$1.1 & 65.5$\pm$4.2 & 76.0$\pm$1.6 & \underline{77.6$\pm$0.8} & 80.5$\pm$2.7 & 71.03\\
\midrule
GTOT-Tuning~(Ours) & \textbf{70.0$\pm$2.3}$\uparrow_{2.0} $ & {75.6$\pm$0.7}$\downarrow_{0.1} $ & \textbf{64.0$\pm$0.3}$\uparrow_{0.1}$ & \textbf{63.5$\pm$0.6}$\uparrow_{2.6}$ & \underline{72.0$\pm$5.4}$\uparrow_{6.1}$ & \underline{80.0$\pm$1.8}$\uparrow_{4.2}$ & \textbf{78.2$\pm$0.7}$\uparrow_{0.9}$ & \textbf{83.4$\pm$1.9}$\uparrow_{3.8}$ &  \textbf{73.34}$\uparrow_{2.49}$ \\
\bottomrule
    \end{tabular}
    }
    \vspace{-2mm}
    \label{tab:molecular}    
\end{table*}

\begin{table*}[th]
\def\p{$\pm$} 
\setlength\tabcolsep{4pt} 
\centering
\vspace{-0mm}
\caption{Test ROC-AUC (\%) of GIN(supervised\_contexpred) on downstream molecular property prediction benchmarks.}
\vspace{-3mm}
\scalebox{0.74}{
\begin{tabular}{l|llllllll|c}
\toprule 
Methods & BBBP & Tox21 & Toxcast& SIDER & ClinTox & MUV & HIV & BACE & \textbf{Average} \\
\midrule
Fine-Tuning~(baseline) & 68.7$\pm$1.3 & {78.1\p 0.6} & 65.7\p 0.6 & 62.7\p 0.8 & 72.6\p 1.5 & 81.3\p 2.1 & 79.9\p 0.7 & 84.5\p 0.7 & 74.19 \\
\midrule
L2\_SP~\cite{xuhong2018explicit} & 68.5$\pm$1.0 & \underline{78.7$\pm$0.3} & 65.7$\pm$0.4 &	\textbf{63.8$\pm$0.3} & 71.8$\pm$1.6 & 85.0$\pm$1.1 & 77.5$\pm$0.9 & \underline{84.5$\pm$0.9} &74.44 \\
 DELTA\cite{li2018delta} & 68.4$\pm$1.2 & 77.9$\pm$0.2 & 65.6$\pm$0.2 & 62.9$\pm$0.8 & 72.7$\pm$1.9 & \textbf{85.9$\pm$1.3} & 75.6$\pm$0.4 & 79.0$\pm$1.1 & 73.50 \\
 Feature(DELTA w/o ATT) & 68.6$\pm$0.9 & 77.9$\pm$0.2 & 65.7$\pm$0.2 & 63.0$\pm$0.6 & 72.7$\pm$1.5 & \underline{85.6$\pm$1.0} & 75.7$\pm$0.3 & 78.4$\pm$0.7 & 73.45 \\
BSS\cite{chen2019catastrophic} & \underline{70.0$\pm$1.0} & 78.3$\pm$0.4 &  65.8$\pm$0.3 & 62.8$\pm$0.6 & \underline{73.7$\pm$1.3} & 78.6$\pm$2.1 & 79.9$\pm$1.4 & 84.2$\pm$1.0 & 74.16 \\
StochNorm~\cite{kou2020stochastic} & 69.8$\pm$0.9 & 78.4$\pm$0.3 & \underline{66.1$\pm$0.4} & 62.2$\pm$0.7 & 73.2$\pm$2.1 & 82.5$\pm$2.6 & \underline{80.2$\pm$0.7} & 84.2$\pm$2.3 & 74.58 \\
\midrule
GTOT-Tuning~(Ours) & \textbf{71.5$\pm$0.8}$\uparrow_{2.8} $ & \textbf{78.6$\pm$0.3}$\uparrow_{0.5} $ & \textbf{66.6$\pm$0.4}$\uparrow_{0.9}$ & \underline{63.3$\pm$0.6}$\uparrow_{0.6}$ & \textbf{77.9$\pm$3.2}$\uparrow_{5.3}$ & 85.0$\pm$0.9$\uparrow_{3.7}$ & \textbf{81.1$\pm$0.5}$\uparrow_{1.2}$ & \textbf{85.3$\pm$1.5}$\uparrow_{0.8}$ &  \textbf{76.16}$\uparrow_{1.97}$ \\
\bottomrule
    \end{tabular}
    }
    \vspace{-3mm}
    \label{tab:sup_GIN_molecular}    
\end{table*}

\subsection{GTOT Regularizer}
Given the node embeddings $\{\x_i^S\}_{i=1}^{|\V|}$ and $\{\x_i^T\}_{i=1}^{|\V|}$ extracted from pre-trained GNN and fine-tuned GNN message passing period, respectively, we calculate the \textbf{cosine dissimilarity} $\C_{ij}=\frac{1}{2}(1-\cos(\x^S_i,\x^T_j))$ as the cost matrix of MWD. Indeed, cosine dissimilarity is a popular choice that used in many OT application~\citep{chen2020graph,xu2020reliable}. 

Intuitively, in most cases, the more (geodesic) distant two vertices in a graph are, the less similar their features are. This implies that when the MWD is used for fine-tuning, the adjacency of the graph should be taken into account, rather than summing all pairwise distances of the cost matrix. Therefore, we set the mask matrix as the adjacency matrix~$\A$ (with self-loop) here, based on the assumption of 1-hop dependence, i.e., the vertices in the target graph are assumed to be related only to the vertices within 1-hop of the corresponding vertices in the source graph~(Fig.~\ref{fig:maskedWD}). This assumption is reasonable because the neighboring node embeddings extracted through the (pretrained) GNN are somewhat similar~\cite{li2018deeper}, which also reveals that considering only the node-to-node distance but without the neighbors is sub-optimal. We call this MOT with graph topology as \textbf{Graph Topology induced Optimal Transport(GTOT)}.
With marginal distributions being uniform, the \textit{GTOT regularizer} 
is formally defined as
\begin{align}
\label{Eq:gtot}
\small
 \mbf{L}_{mw}(\A, \q,\q)= \min\limits_{\P\in \mbf{U}(\A,\q,\q)}\left< \AP,\C\right> \text{(GTOT regularizer)}
\end{align}
where $\q$ is defined as uniform distribution $\mbf{1}_{|\V|}/|\V|$. Noted that the inner product in Eq. \eqref{Eq:gtot} can be rewritten as $\sum_{i}\sum_{j \in \mcal{N}(i)\bigcup\{i\}}\P_{ij}\C_{ij}$, where $\mcal{N}(i)$ denotes the set of neighbors of node $i$. 
We have tried to incorporate the graph structure into the OT using different methods, such as adding a large positive number to the cost at non-adjacent positions directly, but it is not as concise as our method and is tricky to avoid the summation of non-adjacent vertices due to the challenges posed by large numerical computations.
Moreover, our method has the potential to use sparse matrix acceleration when the adjacency matrix is sparse.
Our method is easy to extend to i) the weighted graph by element-wise multiplying the edge-weighted matrix $\W$ with cost matrix, i.e. $\left< \AP,\W\odot \C\right>$ or ii) the k-hop dependence assumption by replacing $\A$ with $\A^k$ or $g(\A)$, where $g$ is a polynomial function.

Similarly, using MOT we can define the MGWD-based Regularizer for regularizing  the representations on edge-level. We defer the details  to Appendix \appref{sec:MGWD_regularizer} since we'd like to focus  on node-level representation regularization with MWD.

\subsection{GTOT-Tuning Framework}
Unlike the OT representation regularizer proposed by \citet{li2020representation} which calculates the Wasserstein distance between mini-batch samples of training data, GTOT regularizer in our framework focus on the single sample, that is, the GTOT distance between the node embeddings of one source graph and the corresponding target graph. Considering the GTOT distance between
nodes allows for knowledge transfer at the node level. This makes the fine-tuned model able to output representations that are more appropriate for the downstream task, i.e., node representations that are as identical as possible to these output from the pre-trained model, but with specific differences in the graph-level representations.

\paragraph{Overall Objective.}
Given $N$ training samples  $\{(G_1,y_1),\cdots,(G_N,y_N)\}$, the overall objective of GTOT-Tuning is to minimize the following loss: 
\begin{align}
    \small
    \mcal{L}=\frac{1}{N}\sum_{i=1}^{N}l(f,G_i,y_i) \label{eq:objective_function}
\end{align}
where $l(f,G_i,y_i):=\phi(f(G_i),y_i)+ 
    \lambda \mbf{L}_{mw}(\A^{(i)},\q^{(i)},\q^{(i)})$, $f$ denotes a given GNN backbone, $\lambda$ is a hyper-parameter for balancing the regularization with the main loss function, and $\phi(\cdot)$ is Cross Entropy loss function.

\section{Theoretical Analysis}
We provide some theoretical analysis for GTOT-Tuning.

\vspace{-1mm}
\paragraph{Related to Graph Laplacian.}
Given a graph signal $s\in \R^{n\xx1}$, if one defines $\C_{ij}:=(s_i-s_j)^2$, then $\L_{mw}=\min_{\P\in \mbf{U}(\A,\a,\b)}\sum_{ij}\P_{ij}\A_{ij}(s_i-s_j)^2$. As we know, $2s^T\L_as = \sum_{ij}\A_{ij}(s_i-s_j)^2$, where $\L_a=\D-\A$ is the Laplacian matrix and $\D$ is the degree diagonal matrix. Therefore, our distance can be viewed as giving a smooth value of the graph signal with topology optimization.
\vspace{-1mm}
\paragraph{Algorithm Stability and Generalization Bound.} We analyze the generalization bound of GTOT-Tuning and expect to find the key factors that affect its generalization ability. We first give the uniform stability below.
\begin{restatable}[Uniform stability for GTOT-Tuning]{lemma}{restadeLemOne}
\label{le:uniform_stability}
    Let $S:=\{z_1=(G_1,y_1),z_2=(G_2,y_2),\cdots,z_{i-1}=(G_{i-1},y_{i-1}),z_i=(G_i,y_i),z_{i+1}=(G_{i+1},y_{i+1}),\cdots,z_N=(G_N,y_N)\}$ be a training set with $N$ graphs, $S^i:=\{G_1,G_2,...,G_{i-1},G_i',G_{i+1},...,G_N\}$ be the training set where graph $i$ has  been replaced. Assume that the number of vertices $|\V_{G_j}|\leq B$ for all $j$ and $0\leq\phi(f_S,z)\leq M$, 
    then
\begin{align}
\small
    |l(f_S,z)-l(f_{S^i},z)|\leq 2M+\lambda \sqrt{B}
\end{align}
where $\lambda$ is the hyper-parameter used in Eq. \eqref{eq:objective_function}. 

\end{restatable}
Based on Lemma \ref{le:uniform_stability} and following the conclusion of \cite{bousquet2002stability}, the generalization error bound of GTOT-Tuning is obtained as follows.
\vspace{-1mm}
\begin{proposition}
\label{pro:generalization_bound}
    Assume that a GNN with GTOT regularization satisfies $0\leq l(f_S,z)\leq \mcal{Q}$. For any $\delta\in (0,1)$, the following bound holds with probability at least $1-\delta$ over the random draw of the sample $S$,
    \begin{align}
    \small
        R(f_S)\leq & R_m(f_S)+ 4M+2\lambda \sqrt{B} \notag \\
        &+ (8N M+4N\lambda\sqrt{B} +\mcal{Q})\sqrt{\frac{\ln{1/\delta}}{2N}}
    \end{align}
\end{proposition}
\noindent where  $R(f_S)$ denotes the generalization error and $R_m(f_S)$ represents empirical error. Proof is deferred to Appendix. 
This result shows that the  generalization bound of GNN with GTOT regularizer is affected by the maximum number of vertices~($B$) in the training dataset.

\section{Experiments}
We conduct experiments on graph classification tasks to evaluate our methods.
\vspace{-2mm}

\subsection{Comparison of Various Fine-tuning Strategies.}
\paragraph{Settings.}
We reuse two pretrained models released by \cite{hu2020strategies}(\url{https://github.com/snap-stanford/pretrain-gnns}) as backbones: GIN~(contextpred)~\cite{xu2018powerful}, which is only pretrained via self-supervised task \textit{Context Prediction}, and GIN~(supervised\_contextpred), an  architecture that is pretrained by \textit{Context  Prediction + Graph Level} multi-task supervised strategy. Both networks are pre-trained on the Chemistry dataset (with 2 million molecules).
In addition, eight binary classification datasets in MoleculeNet \cite{wu2018moleculenet} serve as downstream tasks for evaluating the fine-tuning strategies, where the scaffold split scheme is used for dataset split.
More details can be found in Appendix.

\noindent\textbf{Baselines.} 
Since we have not found related works about fine-tuning GNNs, we extend several typical baselines tailored for Convolutional networks to GNNs, including L2\_SP~\cite{xuhong2018explicit}, DELTA~\citet{li2018delta}, BSS~\citet{chen2019catastrophic}, SotchNorm~\cite{kou2020stochastic}.

\noindent\textbf{Results.}
The results with different fine-tuning strategies are shown in Table \ref{tab:molecular},\ref{tab:sup_GIN_molecular}. 
Observation (1): GTOT-Tuning gains a competitive performance on different datasets and outperforms other methods on average.  Observation (2): Weights regularization (L2\_SP) can't obtain improvement on pure self-supervised tasks. This implies L2\_SP may require the pretrained task to be similar to the downstream task. Fortunately, our method can consistently boost the performance of both supervised and self-supervised pretrained models.
Observation (3): The performance of the Euclidean distance regularization (Features(DELTA w/o ATT)) is worse than vanilla fine-tuning, indicating that directly using the node representation regularization may cause negative transfer.

\begin{figure}[ht]
    \centering
    \vspace{-3mm}
    \subfigure{
    \begin{minipage}[s]{0.47\linewidth}
    \centering
    \includegraphics[width=1\linewidth]{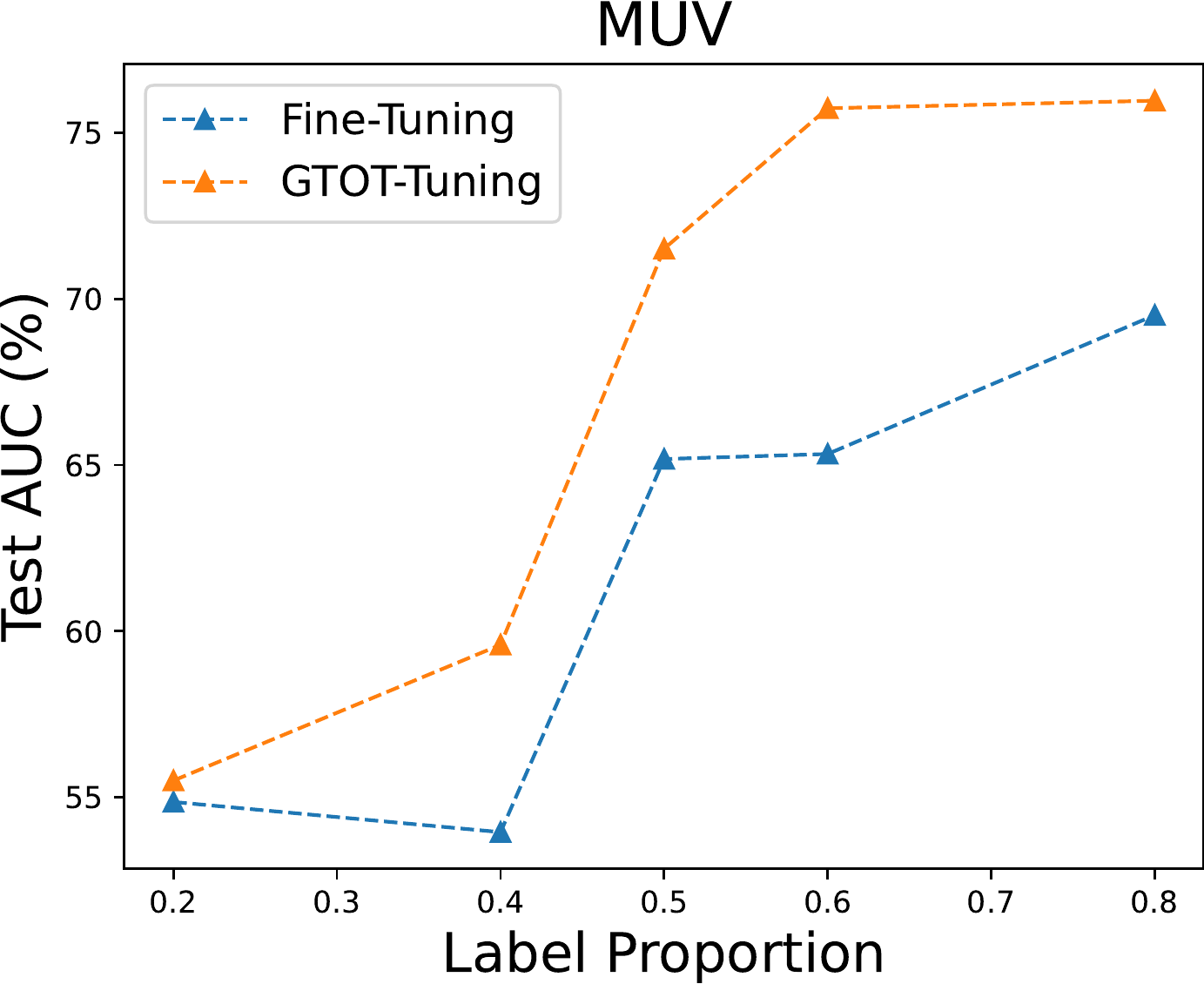}
    \end{minipage}
    }
    \subfigure{
    \begin{minipage}[s]{0.47\linewidth}
    \centering
    \includegraphics[width=1\linewidth]{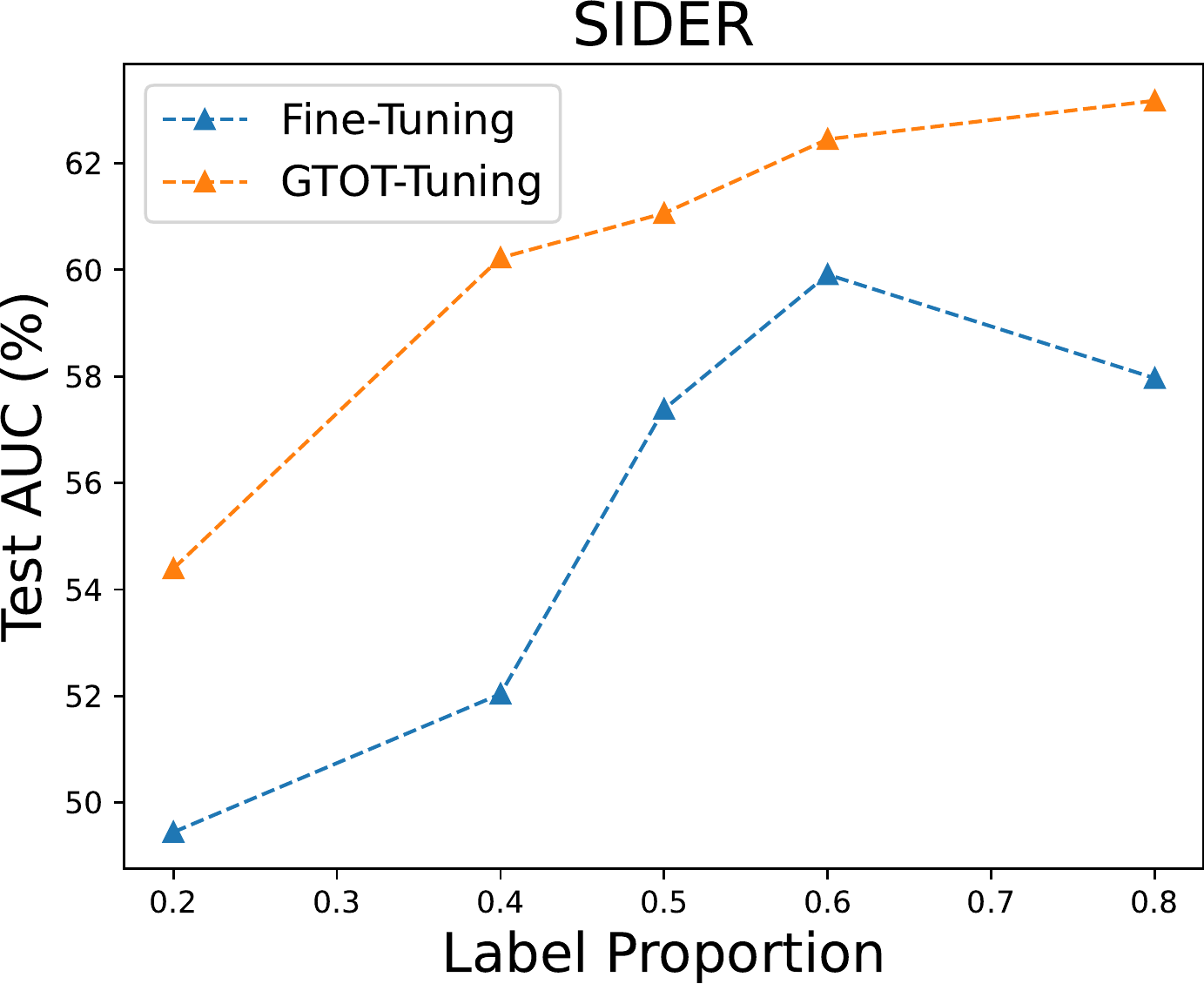}
    \end{minipage}
    }
    \vspace{-3mm}
     \caption{Test AUC with different proportion of labeled data. }
    \label{fig:label_radio}
        \vspace{-4mm}
\end{figure}
\subsection{Ablation Studies}
\textbf{Effects of the Mask Matrix.}
We conduct the experiments on GNNs fine-tuning with GTOT regularizer to verify the efficiency of the introduced adjacency matrix. 
The results shown in Table \ref{tab:ab_ot} suggest that when using adjacency matrix as mask matrix, the performance on most downstream tasks will be better than using classic WD directly.  
Besides, the competitive performance when the mask matrix is identity also implies that we can select possible pre-designed mask matrices, such as polynomials of $\A$, for specific downstream tasks. This also indicates that our MOT can be flexibly used for fine-tuning.
\begin{table}[th]
\def\p{$\pm$} 
\centering
\vspace{-2mm}
\caption{Test ROC-AUC (\%) of GIN(contextpred) on downstream tasks. (The parentheses indicate the masked matrix. MWD(A) is GTOT distance.)}
\setlength\tabcolsep{6pt}
\vspace{-3mm}
\scalebox{0.85}{
\begin{tabular}{l|cccc}
\toprule 
Methods & BBBP & Tox21 & Toxcast& SIDER   \\
\midrule
\# BPTs & 1 & 12 & 617 & 27 \\
\# $\mathcal{A}$\_distance &  1.57 & 1.56 &1.56& 1.41 \\
\midrule
w/o MWD & 68.7$\pm$3.4 & 75.9$\pm$0.5 & 63.1$\pm$0.6 & 60.2$\pm$0.9  \\
w/ MWD$(\mbf{1}_{n\times n})$ & 66.2$\pm$3.3 & 75.3$\pm$0.9 & 63.6$\pm$0.6 & 62.7$\pm$0.7 \\
w/ MWD$(\A)$ & \textbf{69.6$\pm$2.6} & \textbf{75.7$\pm$0.5}  & {63.8$\pm$0.4}  & 63.5$\pm$0.6 \\
w/ MWD$(\mbf{I})$ &  68.6$\pm$3.5 & 75.4$\pm$0.7 & \textbf{64.1$\pm$0.3} & \textbf{63.7$\pm$0.5} \\
\midrule
 Methods & ClinTox & MUV & HIV & BACE \\
\midrule
\# BPTs & 2 & 17 & 1 & 1 \\
 \# $\mathcal{A}$\_distance & 1.41 & 1.19 & 1.65 & 1.65 \\
 \midrule
 w/o MWD &  69.5$\pm$5.0 &69.5$\pm$1.3  & 78.2$\pm$1.2 & 82.5$\pm$1.7 \\
w/ MWD$(\mbf{1}_{n\times n})$ &  69.5$\pm$5.0 & 74.3$\pm$1.3 & 78.2$\pm$0.8 & \textbf{83.7$\pm$1.9} \\
w/  MWD$(\A)$ &  \textbf{70.9$\pm$5.8} &  \textbf{80.7$\pm$0.6} & \textbf{78.5$\pm$1.5}  & 83.1$\pm$1.9\\
 w/ MWD$(\mbf{I})$ &  69.7$\pm$4.0 & 80.2$\pm$0.9 & 78.3$\pm$1.3  & 82.6$\pm$2.5 \\
\bottomrule
    \end{tabular}
    }
    \vspace{-3mm}
    \label{tab:ab_ot}    
\end{table}

\noindent\textbf{Effects of Different Proportion of Labeled Data.} We also investigate the performance of the proposed method with different proportions of labeled data on MUV and SIDER datasets. As illustrated in Fig. \ref{fig:label_radio},
MWD outperforms the baseline~(vanilla Fine-Tuning) given different amounts of labeled data consistently, indicating the generalization of our method.

\begin{figure}[htbp]
\vspace{-2mm}
\centering
    \subfigure{
    \begin{minipage}[s]{0.98\linewidth}
    \centering
    \includegraphics[width=1\linewidth]{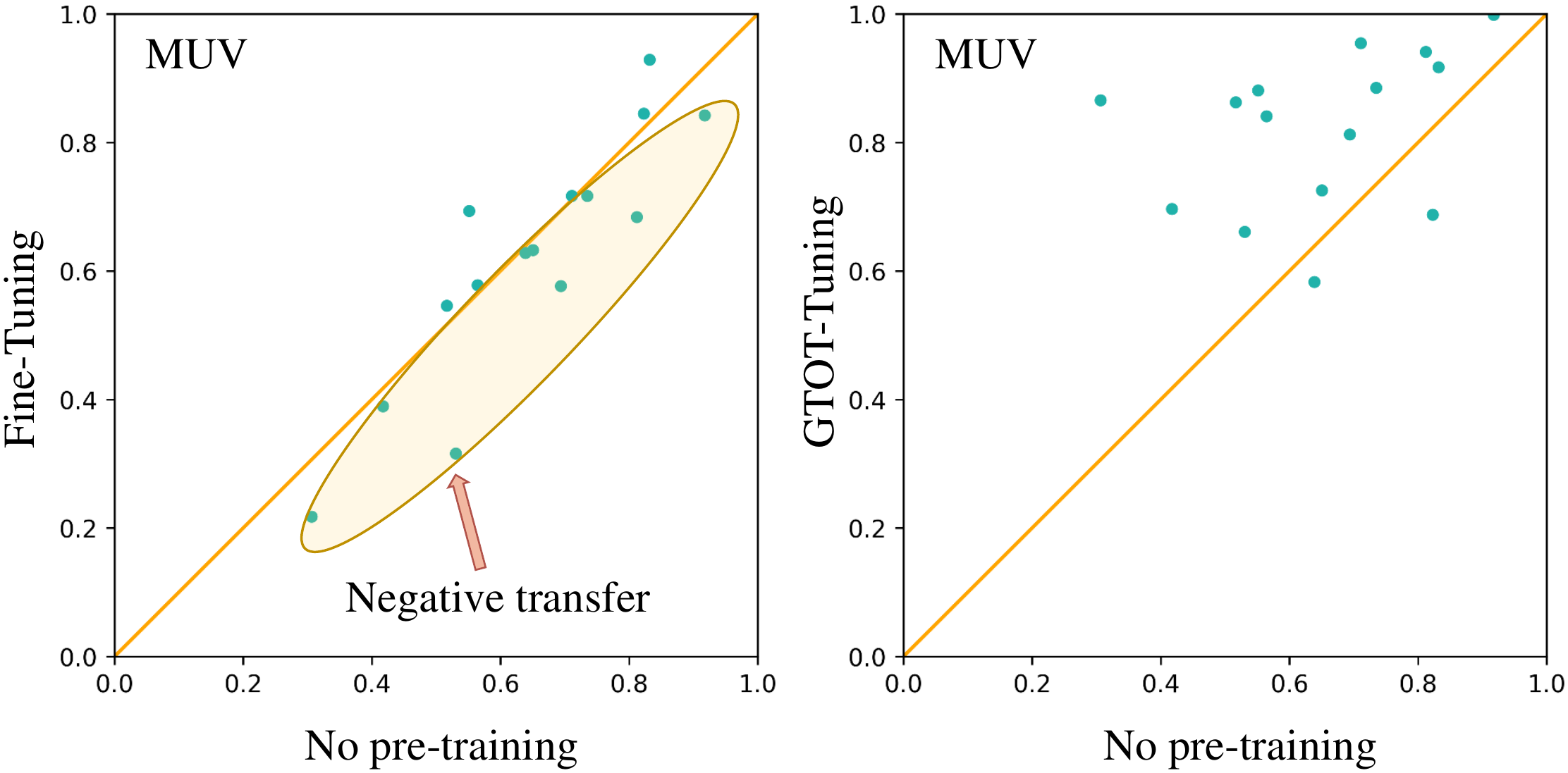}
    \end{minipage}%
    }%
\centering
\vspace{-4mm}
\caption{Scatter plot
comparisons of ROC-AUC scores for a pair of fine-tuning strategies on the multi-task datasets (MUV). Each point represents a particular downstream task. There are many downstream tasks where vanilla fine-tuning performs worse than non-pre-trained model, indicating negative transfer. Meanwhile, negative transfer is alleviated when the GTOT regularizer is used. }
\vspace{-3mm}
\label{fig:negative_transfer}
\end{figure}
\subsection{Why GTOT-Tuning Works}\label{sec:why_work}
\begin{figure}[htbp]
\centering
\vspace{-3mm}
    \subfigure{
    \begin{minipage}[s]{0.48\linewidth}
    \centering
    \includegraphics[width=1\linewidth]{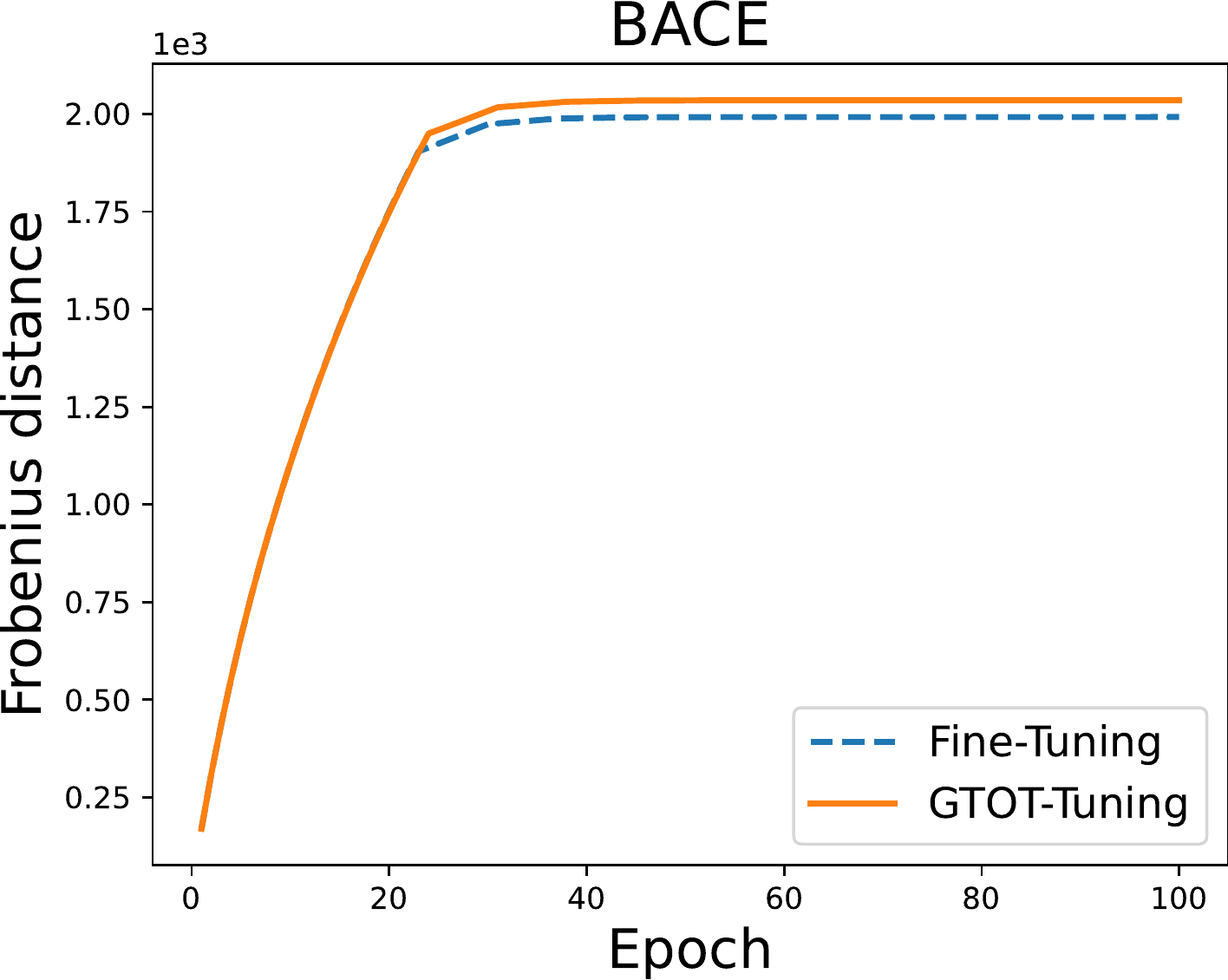}
    \end{minipage}%
    }%
    \subfigure{
    \begin{minipage}[s]{0.48\linewidth}
    \centering
    \includegraphics[width=1\linewidth]{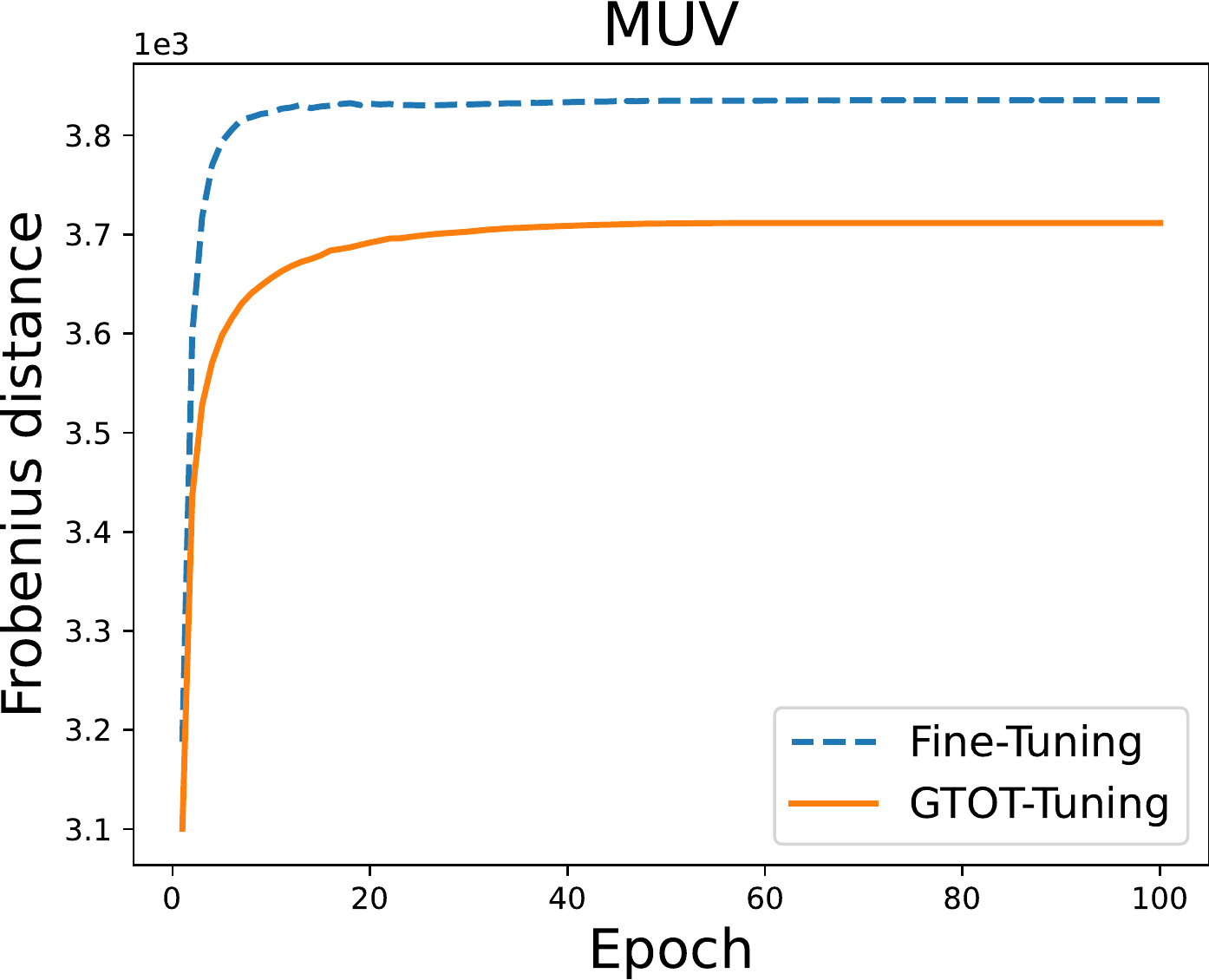}
  
    \end{minipage}%
    }%
\vspace{-3mm}
\caption{Weights distance between pre-trained initialization weights and fine-tuned weights. Benefit from the soft alignment of MOT, the GTOT regularizer is able to implicitly adjust the distance of the weights according to the downstream task.}
\vspace{-3mm}
\label{fig:adaptive_domain_gap}
\end{figure}

a) \textbf{Adaptive Adjustment of Model Weights to Downstream Tasks.}
The $\mathcal{A}$-distance~\cite{ben2007analysis}, $d_{A}(D^S, D^T
) = 2(1-2\xi(h))$ is adopted to measure the discrepancy between the pretrained and finetuned domains, i.e. the domain gap between pre-trained data $D^S$ and fine-tuned data $D^T$. $\xi(h)$ is the error of a linear SVM classifier $h$ discriminating the two domains~\cite{xu2020reliable}. We calculate the $d_{A}$ with representations extracted by GIN~(contextpred) as input and show the results  in Table \ref{tab:ab_ot}.
From Fig.  \ref{fig:adaptive_domain_gap},
it can be observed that GTOT-Tuning
constrains the weights distance between fine-tuned and pre-trained model when domain gap is relatively small~(MUV).
Conversely, when the domain gap is large~(BACE), our method does not necessarily increase the distance between weights, but rather increases the distance between weights.
This reveals that GOT-Tuning can adaptively and implicitly adjust the distance between fine-tuned weights and pre-trained weights according to the downstream task, yielding powerful fine-tuned models.

\noindent b) \textbf{Mitigating Negative Transfer under Multi-task.} Fig.  \ref{fig:negative_transfer} shows that GTOT-Tuning boosts the performance of most tasks on multi-task datasets, demonstrating the ability of our method in mitigating negative transfer.

\noindent Due to space limitations, we defer additional experimental results on GCN~(contextpred) backbone, sensitivity analysis, runtimes to Appendix~\ref{subsec:additional_experiments}, \ref{subsec:Hyper-parameter_ Sensitivity}, \ref{subsec:runtimes}, respectively. Code would be available
at \url{https://github.com/youjibiying/GTOT-Tuning}.
\vspace{-2mm}

\section{Discussions and Future Work}
Despite the good results, there are some aspects which worth further explorations in the future: i)
The MOT-based methods require relatively high computational cost, which calls for designing a more efficient algorithm to solve it. 
ii) MWD has the potential to perform even better when used in conjunction with MGWD, which may require a more suitable combination by design. iii) The optimal transport plan obtained by MOT can be used to design new message passing schemes in graph. iv) The proposed method can be potentially extended for more challenging settings that needs advanced knowledge transfer techniques, such as the graph out of distribution learning \cite{2022arXiv220109637J}.  v) Other applications in graph knowledge distillation or MOT barycenters can be envisioned.

\section*{Acknowledgments}
 The authors would like to thank Ximei Wang ,Guoji Fu and Guanzi Chen for their sincere and selfless help.


\bibliographystyle{named}
\bibliography{0_main.bib}

\clearpage
\onecolumn
\appendix
    \begin{center}
    \Large
    \textbf{Appendix}
     \\[20pt]
    \end{center}
    
\etocdepthtag.toc{mtappendix}
\etocsettagdepth{mtchapter}{none}
\etocsettagdepth{mtappendix}{subsection}
\tableofcontents

\section{Derivations of Algorithm \ref{alg:gtot}}
\subsection{Masked Optimal Transport Problem.}

\restadefMWD*

\restadeproMWD*

\paragraph{Proof.}
Introducing two dual variables  $\mbf{f}\in {\R}^{n},\mathbf{g}\in {\R}^{m}$ 
for the two marginal constraints, respectively,
the Lagrangian of Eq. \eqref{eq:MWD} with entropic regularization  is

\begin{align}
     L(\P,\f,\g)=\left<\MP,\C\right>-\epsilon\H(\MP)-\left<\f,\MP\mathbf{1}_m-\a\right>-\left<\g,(\MP)^{\top}\mathbf{1}_n-\b\right>.
\end{align}

Using first order conditions, we get
\begin{align*}
    \frac{\partial L(\P,\f,\g)}{\partial \P_{ij}}=\M_{ij}\C_{ij} + \epsilon\M_{ij}\log(\M_{ij}\P_{ij})-\f_i\M_{ij}-\g_j\M_{ij} = 0.
\end{align*}
That is
\begin{align*}
    \exp(\M\odot \log(\MP))=\exp(diag(\f)\M/\epsilon)\odot\exp(-\C\odot\M/\epsilon)\odot\exp(\M diag(\g)/\epsilon)
\end{align*}
where $\exp(\X):=\exp(\X_{ij})$, $\exp(\x):=\exp(\x_{i})$ for any matrix $\X$ or vector $\x$, respectively. Let $\bar{\M}_{ij}:=1-\M_{ij}$.
Note that $\forall ~\M\in\{0,1\}^{n\times m}, \forall~\X\in\R^{n\times m}$ , we have $\exp(\M\odot\X) = \M\odot \exp(\X)+\bar{\M}$. Therefore, $\exp(diag(\f)\M/\epsilon)=\M\odot(\exp(\f/\epsilon)\mbf{1}_n^{\top}) +\bar{\M}$. 
Then we have
\begin{align*}
    \M\odot\exp(\log(\MP)) & +\bar{\M} = \exp(diag(\f)\M/\epsilon)\odot\exp(-\C\odot\M/\epsilon)\odot\exp(\M diag(\g)/\epsilon) 
\end{align*}
Further simplification yields
\begin{align*}
    \MP &= \exp(diag(\f)\M/\epsilon)\odot\exp(-\C\odot\M/\epsilon)\odot\exp(\M diag(\g)/\epsilon)-\bar{\M} \\
    & = (\M\odot(\exp(\f/\epsilon)\mbf{1}_n^{\top}) +\bar{\M}) \odot \exp(-\C\odot\M/\epsilon) \odot (\M\odot(\mbf{1}_m\exp(\g/\epsilon)^{\top}) +\bar{\M}) -\bar{\M}\\
    & = \M\odot(\exp(\f/\epsilon)\mbf{1}_n^{\top})  \odot \exp(-\C\odot\M/\epsilon) \odot (\M\odot(\mbf{1}_m\exp(\g/\epsilon)^{\top}) 
     +\bar{\M}\odot \exp(-\C\odot\M/\epsilon)  \odot\bar{\M} -\bar{\M}\\
    &=(\exp(\f/\epsilon)\mbf{1}_n^{\top}) \odot \K \odot \M\odot (\mbf{1}_m\exp(\g/\epsilon)^{\top}).
\end{align*}
From the constrain $\P\odot(\mbf{1}_{n\xx m}-\M)=\mbf{0}_{n\xx m}$ in $\U(\M,\a,\b)$, we have $\P_{ij}=0$ if $\M_{ij}=0$. To sum up, the P can be uniformly written as
\begin{align}
    \label{eq:pkfg}
    \P= (\exp(\f/\epsilon)\mbf{1}_n^{\top}) \odot \K \odot \M\odot (\mbf{1}_m\exp(\g/\epsilon)^{\top}) 
\end{align}
Replacing $\exp(\f/\epsilon)$ and $\exp(\g/\epsilon)$  with nonnegative vectors $\u,\v$, respectively
, we get
\begin{align}
    \P_{ij} = \u_i\M_{ij}\K_{ij}\v_j.
\end{align}

\subsection{Masked Sinkhorn Iterations}\label{subsec:MSI}
\begin{proposition}[Masked Sinkhorn Iterations]
The masked optimal transport problem can be solved iteratively. These two updates define masked Sinkhorn's algorithm,
\begin{align}
\label{eq:mask_sinkhorn_iterations}
    \u^{(k+1)}:=\frac{\a}{(\M\odot\K)\v^{(k)}} \quad \text{ and }\quad \v^{(k+1)}:=\frac{\b}{(\M\odot\K)^{\top}\u^{(k+1)}}
\end{align}
initialized with an arbitrary positive vector $\v^{(0)}=\mathbbm{1}_m$.
\end{proposition}

\paragraph{Proof:}
Substituting Eq. \eqref{eq:solver} into constraints $(\MP) \mbf{1}_m = \a, (\MP)^{\top}\mbf{1}_n = \b$, we have
\begin{align*}
    \sum_j \P_{ij}\M_{ij}=\sum_j \u_i\M_{ij}\K_{ij}\v_j=\a_i \\
    \u_i = \frac{\a_i}{\sum_j\M_{ij}\K_{ij}\v_j}
\end{align*}
Similarly, 
\begin{align*}
    \v_j =\frac{\b_j}{\sum_i\u_i\M_{ij}\K_{ij}}
\end{align*}

\begin{remark}[Complexity of Masked Sinkhorn Algorithm]\label{appPro:complexity}
Assume that $n = m$ in Definition 2 for the sake of simplicity.
Let $\hat{\P}$ denote a approximate solution satisfying
$\left<\hat{\P}, \C \right> \leq \L_{mw}(\M,\a, \b)+\tau $. When  $\epsilon = \frac{\tau}{4log(n)}$, from \appcitet{altschuler2017near}, we know the time complexity of the classical Sinkhorn iterations to get $ \hat{\P}$ is $O(n^2\|\C\|^{2}_{\infty} \tau^{-3}(\tau \log(s)+\|\C\|_{\infty}\log(n)) )$, where $\|\C\|_\infty := \max_{ij}\C_{ij}$ and $s:=\sum_{ij}(\K)_{ij}$. In MOT, the $\K$ in OT can be replaced with $\M\odot\K$, so the $s$ can be replaced with $s'=\sum_{ij}\M_{ij}\K_{ij} \leq s$. Hence our Masked optimal transport complexity is $O(n^2\|\C\|^{2}_{\infty} \tau^{-3}(\tau \log(s')+\|\C\|_{\infty}\log(n)) ) \leq O(n^2\|\C\|^{2}_{\infty} \tau^{-3}(\tau \log(s)+\|\C\|_{\infty}\log(n)) )$. This means that 
the complexity of our method is smaller than the typical OT Sinkhorn algorithm.
\end{remark}

\subsection{Log-domain Masked Sinkhorn}\label{subsec:LD_Masked_Sinkhorn}
\begin{proposition}
 The masked optimal transport problem \eqref{eq:MWD} is equivalent to 
\begin{align}
\label{eq:dual_problem}
   \L^{\epsilon}_{\M\C} =  \max\limits_{\f\in \R^{n},\g\in\R^{m} }\left< \f,\a\right> +\left< \g,\b\right> - \epsilon\left<e^{\f/\epsilon}, (\M\odot\K)e^{\g/\epsilon}\right>
\end{align}
where $(\f,\g)$ are linked to $(\u,\v)$ that appears in \eqref{eq:solver} via $(\u,\v)=(e^{\f/\epsilon},e^{\g/\epsilon})$.
\end{proposition}

\paragraph{Proof.} From Eq. \eqref{eq:pkfg}, we know the optimal primal solution $\P$ and dual multipliers $\f$ and $\g$ 
satisfy the following equation:
\begin{align*}
\P= (\exp(\f/\epsilon)\mbf{1}_n^{\top}) \odot \K \odot \M\odot (\mbf{1}_m\exp(\g/\epsilon)^{\top}) =\M\odot e^{(\f\mbf{1}_n^{\top}/\epsilon)} \odot e^{-\C/\epsilon} \odot e^{(\mbf{1}_m\g^{\top}/\epsilon)}
\end{align*}
Substituting the optimal $\P$ into the problem \eqref{eq:MWD}, we obtain the equivalent Lagrangian dual function as 
\begin{align}
\label{eq:Lagrangian_dual_func_fg}
    \left<e^{\f/\epsilon},(\K\odot\M\odot\C)e^{\g/\epsilon})\right>-\epsilon\H(\M \odot e^{(\f\mbf{1}_n^{\top}/\epsilon)} \odot e^{-\C/\epsilon} \odot e^{(\mbf{1}_m\g^{\top}/\epsilon)})\propto (\f,\g)
\end{align}
Note that

\begin{align*}
-\epsilon\H(\MP) & = \left<\MP,\epsilon\log(\MP)-\epsilon \mbf{1}_{n\times m}\right> \\
& =\left<\MP,\epsilon\log(\M\odot e^{(\f\mbf{1}_n^{\top}/\epsilon)} \odot e^{-\C/\epsilon} \odot e^{(\mbf{1}_m\g^{\top}/\epsilon)}) -\epsilon \mbf{1}_{n\times m}\right> \\
& =\left<\MP , \f\mbf{1}_n^{\top}-\C + \mbf{1}_m\g^{\top}  -\epsilon \mbf{1}_{n\times m} \right> \\
&= -\left<\MP ,\C\right> + \left<\MP ,\f\mbf{1}_n^{\top}\right> + \left<\MP ,\mbf{1}_m\g^{\top}\right> -\epsilon\left< \MP,\mbf{1}_{n\times m} \right>\\
&= -\left<\MP ,\C\right> + \left< \a,\f\right> + \left< \b,\g\right> -\epsilon\left< \MP,\mbf{1}_{n\times m} \right>\\
&= -\left<\MP ,\C\right> + \left< \a,\f\right> + \left< \b,\g\right> -\epsilon\left<\M\odot e^{(\f\mbf{1}_n^{\top}/\epsilon)} \odot e^{-\C/\epsilon} \odot e^{(\mbf{1}_m\g^{\top}/\epsilon)} ,\mbf{1}_{n\times m}\right> \\
&= -\left<\MP ,\C\right> + \left< \a,\f\right> + \left< \b,\g\right> -\epsilon\left<e^{\f/\epsilon},(\K\odot\M)e^{\g/\epsilon})\right>. \\
\end{align*}
The last equation 
comes from that
\begin{align*}
    \left<\M\odot e^{(\f\mbf{1}_n^{\top}/\epsilon)} \odot e^{-\C/\epsilon} \odot e^{(\mbf{1}_m\g^{\top}/\epsilon)} ,\mbf{1}_{n\times m}\right>  &= \sum_{ij}e^{\f_i}\M_{ij}\K_{ij}e^{\g_j} \\
    &= \sum_i e^{\f_i} \sum_{j}\M_{ij}\K_{ij}e^{\g_j}=\left<e^{\f/\epsilon},(\K\odot\M)e^{\g/\epsilon})\right>.
\end{align*}

 Putting the result back to Eq. \eqref{eq:Lagrangian_dual_func_fg}, we have
 \begin{align*}
     \left< \MP,\C\right>-\epsilon\H(\MP)&= \left< \MP,\C\right>-\left<\MP ,\C\right> + \left< \a,\f\right> + \left< \b,\g\right> -\epsilon\left<e^{\f/\epsilon},(\K\odot\M)e^{\g/\epsilon})\right> \\
     & = \left< \f,\a\right> + \left< \g,\b\right> -\epsilon\left<e^{\f/\epsilon},(\K\odot\M)e^{\g/\epsilon})\right> \propto (\f,\g)
 \end{align*}

\begin{proposition}[Log-domain Masked Sinkhorn]
The Log-domain Masked Sinkhorn is defined as a two-step update
 \begin{align}
     \f^{(k+1)}= & \epsilon\log(\a)-\epsilon\left(
    \log([\M\odot  \exp(-\C+\f^{(k)}\mathbf{1}_m^{\top}+\mathbf{1}_n(\g^{(k)})^{\top})]\mbf{1}_m)\right) + \f^{(k)} \\
    \g^{(k+1)}= &\epsilon \log(\b)-\epsilon \left( 
    \log([\M\odot \exp(-\C+\f^{(k+1)}\mathbf{1}_m^{\top}+\mathbf{1}_n(\g^{(k)})^{\top}]^{\top}\mathbf{1}_n)\right) + \g^{(k)}
 \end{align}
 with initialized $\f^{(0)}=\mbf{0},\g^{(0)}=\mbf{0}$.
\end{proposition}
\paragraph{Proof.} Writing $\mathcal{F}(\f,\g):=\left< \f,\a\right> +\left< \g,\b\right> - \epsilon\left<e^{\f/\epsilon}, (\M\odot\K)e^{\g/\epsilon}\right> $ for the objective of \eqref{eq:dual_problem}. The unconstrained maximization problem \eqref{eq:dual_problem} can be solved by an exact
\textit{block coordinate ascent} strategy
\begin{align}
    \nabla |_{\f} \mathcal{F} &= \a - e^{\f/\epsilon}\odot (\M\odot \K) e^{\g/\epsilon}= 0 \\
    \nabla |_{\g} \mathcal{F} &= \b - e^{\g/\epsilon}\odot (\M\odot \K) e^{\f/\epsilon}= 0
\end{align}
Applying the following updates, with any initialized vector $\g^{(0)}$, we have
\begin{align}
    \f^{(k+1)} &= \epsilon\log(\a)-\epsilon\log((\M\odot\K)e^{\g^{(k)}/\epsilon}) \label{eq:block coordinate ascent1} \\
    \g^{(k+1)} &= \epsilon\log(\b)-\epsilon\log((\M\odot\K)e^{\f^{(k+1)}/\epsilon}) \label{eq:block coordinate ascent2}
\end{align}
Indeed, the updates above are equivalent to the masked Sinkhorn updates in
\eqref{eq:mask_sinkhorn_iterations} by applying the primal-dual relationships
\begin{align*}
    (\f^{(k)},\g^{(k)}) = \epsilon(\log(\u^{(k)}),\log(\v^{(k)}))
\end{align*}
Following the derivation of log-domain Sinkhorn iterations (\appcitet{peyre2020computational},Remark 4.22), we use \textit{soft-min rewriting } to write $\operatorname{min}_\epsilon \mbf{z}$ for the \textit{soft-minimum} of the its coordinates
\begin{align}
    \operatorname{min}_\epsilon = -\epsilon\log\sum_i e^{-\mbf{z}_i/\epsilon}.
\end{align}
Obviously, $\lim_{\epsilon\to 0 }\operatorname{min}_\epsilon \mbf z = \operatorname{min} \mbf z$.  
Based on this notation, we further define the operator that takes an arbitrary given matrix $\A\in\R^{n\times m}$ and outputs a column vector of soft-minimum vales of its columns or rows
\begin{align*}
    \operatorname{Min}_{\epsilon}^{row}(\A)_i & := \operatorname{min}_\epsilon(\A_{i,*}), \forall i \in \{1,2,...,n\} \\
    \operatorname{Min}_{\epsilon}^{col}(\A)_j & := \operatorname{min}_\epsilon(\A_{*,j}),\forall j \in \{1,2,...,m\}
\end{align*}

Using this notation, Eq. \eqref{eq:block coordinate ascent1} \eqref{eq:block coordinate ascent2} can be rewritten as 
\begin{align}
    \f^{(k+1)} &= \epsilon\log(\a)+\operatorname{Min}^{row}_\epsilon (\log(\M)+\C-\mbf{1}_n(\g^{(k)})^{\top}) \\
    \g^{(k+1)} &= \epsilon\log(\b)+\operatorname{Min}^{row}_\epsilon (\log(\M)+\C-\f^{(k)}\mbf{1}_m^{\top})
\end{align}
where we define $\log(0):=-\infty$ and $e^{\log(0)}:=0$ to make the $\log(\M)$ well-defined.
 This \textit{log-sum-exp} trick can avoid underflow for small values of $\epsilon$, when $\operatorname{min}_\epsilon\to \operatorname{min}$ as $\epsilon\to 0 $. Writing $\underline{z} = \operatorname{min}\mbf{z}$, \appcitet{peyre2020computational} suggest evaluating $\operatorname{min}_\epsilon \mbf{z}$ as 
\begin{align}
    \operatorname{min}_\epsilon \mbf{z} = \underline{z} -\epsilon\log\sum_i e^{-(\mbf{z}_i-\underline{z})/\epsilon}
\end{align}
and replacing the $\underline{z}$ with the practical scaling computed previously.
Finally, the stabilized iterations of log-domain masked Sinkhorn can be 
obtained as
\begin{align}
    \f^{(k+1)}&=\operatorname{Min}^{row}_{\epsilon} (\mcal{S}(\f^{(k)},\g^{(k)})+\f^{(k)} +\epsilon\log(\a) \label{eq:block coordinate ascent_Min1}\\
    \g^{(k+1)} &= \operatorname{Min}^{col}_{\epsilon} (\mcal{S}(\f^{(k+1)},\g^{(k)}))+\g^{(k)} +\epsilon\log(\b)  \label{eq:block coordinate_ascent_Min2}
\end{align}
where we defined 
\begin{align}
    \mcal{S}(\f^{(k)},\g^{(k)}) := \log(\M)+\C-\f^{(k)}\mbf{1}_m^{\top}-\mbf{1}_n(\g^{(k)})^{\top}
\end{align}
\textbf{The only difference from the classical log-domain sinkhorn algorithm is that here we have an additional term $\log(\M)$.} Note that when $\M=\mbf{1}_{n\times m}$ , $\log(\M)=\mbf{0}$. Therefore, our masked iterations recover the non-masked iterations and can be seen as an extension of classical results \appcite{peyre2020computational}.

Rewriting the Eq. \eqref{eq:block coordinate ascent_Min1}, Eq. \eqref{eq:block coordinate_ascent_Min2} to explicit forms 
\begin{align}
       \f^{(k+1)}= & \epsilon\log(\a)-\epsilon\left(
    \log([\M\odot  \exp(-\C+\f^{(k)}\mathbf{1}_m^{\top}+\mathbf{1}_n(\g^{(k)})^{\top})]\mbf{1}_m)\right) + \f^{(k)} \\
    \g^{(k+1)}= &\epsilon \log(\b)-\epsilon \left( 
    \log([\M\odot \exp(-\C+\f^{(k+1)}\mathbf{1}_m^{\top}+\mathbf{1}_n(\g^{(k)})^{\top}]^{\top}\mathbf{1}_n)\right) + \g^{(k)}
\end{align}

\subsection{Masked Gromov-Wasserstein Distance}\label{subsec:MGWD_definition}
\begin{restatable}[Masked Gromov-Wasserstein distance]{definition}{restadefMGWD}
 \label{def:MGWD}
 Following the same notation as Definition \ref{def:MWD},
 the masked Gromov-Wasserstein distance~(MGWD) $\mbf{L}_{mgw}(\M,\a,\b) $ is defined as
 \begin{align}
 \small
      \min\limits_{\P\in \mbf{U}(\M,\a,\b)}\sum_{ijkl}L(\C_{ik},\bar{\C}_{jl}) (\MP)_{ij}(\MP)_{kl} \label{eq:MGWD}
\end{align}
where $\C\in\R^{n\times n}$ and $\bar{\C}\in\R^{m\times m}$ are distance matrices that represent the pair-wise point distance within the same space and $L$ is the cost function evaluating intra-domain structural diversity between two pairs of points $\x_i,\x_j$ and $\y_k,\y_l$.
\end{restatable}

\begin{lemma}
(\appcitet{peyre2016gromov} proposition 1)
    \label{le:GWD}
     For any tensor $\mcal{L} = (\mcal{L}_{i,j,k,l})_{i,j,k,l}$ and matrix $(\P_{i,j})_{i,j}$, the tensor matrix multiplication is defined as
     $\mcal{L} \otimes \P :=\sum_{kl}$
     $
     (\mcal{L}_{i,j,k,l} \P_{kl})_{ij}$
     If the loss can be calculated as 
     \begin{align*}
         L(a, b) =
    f_1(a) + f_2(b) - h_1(a)h_2(b)
     \end{align*}
      for functions $(f_1, f_2, h_1, h_2)$, then, for any $\P\in\U(\mbf{1}_{n\times m},\a,\b)$,
     \begin{align}
     \label{eq:tensor_decompose}
         \mcal{L}(\C,\bar{\C})\otimes\P = c_{\C,\bar{\C}} - h_1(\C)\P h_2(\bar{\C})^{\top}
     \end{align}
    where $c_{\C,\bar{\C}}:= f_1(\C)\a\mbf{1}^{\top}_m + \mbf{1}_{n}\b^{\top}f_2(\bar{\C})^{\top} $ is independent of $\P$.
\end{lemma}
\begin{proposition}
\label{pro:GWD}
 Using the same notation as Lemma \ref{le:GWD}, If the loss can be calculated as 
 \begin{align*}
     L(a, b) =
f_1(a) + f_2(b) - h_1(a)h_2(b)
 \end{align*}
  for functions $(f_1, f_2, h_1, h_2)$, then, for any $\P\in\U(\M,\a,\b)$,
 \begin{align}
 \label{eq:GWD_tensor_decompose}
     \mcal{L}(\C,\bar{\C})\otimes(\MP) = c_{\C,\bar{\C}} - h_1(\C)(\MP) h_2(\bar{\C})^{\top}
 \end{align}
where $c_{\C,\bar{\C}}:= f_1(\C)\a\mbf{1}^{\top}_m + \mbf{1}_{n}\b^{\top}f_2(\bar{\C})^{\top} $ is independent of $\P$.
Let $\hat{\C}$ denote $\mcal{L}(\C,\bar{\C})\otimes(\MP)$ as a pseudo-cost matrix, then the solution to Definition \ref{def:MGWD} with entropic regularization $\epsilon\H(\MP)$ is
\begin{align}
\label{eq:GWD_solver}
    \P_{ij} = \u_i\M_{ij}\hat{\K}_{ij}\v_j
\end{align}
where $\hat{\mbf{K}}_{ij}=\exp(-\hat{\C}_{ij}/\epsilon)$ and $(\u,\v)\in \R^{n}_+\times \R^{m}_+$ are two unknown scaling variables .
\end{proposition}

\paragraph{Proof.} Eq. \eqref{eq:MGWD} can be rewritten as
\begin{align*}
   \sum_{ijkl}L(\C_{ik},\bar{\C}_{jl}) (\MP)_{ij}(\MP)_{kl} = \left<\mcal{L}(\C,\bar{\C})\otimes(\MP),\MP\right>
\end{align*}
From Eq. \eqref{eq:tensor_decompose}, for $(\MP)\in \U(\mbf{1}_{n\times m},\a,\b)$, we have 
\begin{align*}
     \mcal{L}(\C,\bar{\C})\otimes(\MP) = c_{\C,\bar{\C}} - h_1(\C)(\MP) h_2(\bar{\C})^{\top}.
 \end{align*}
Note that $(\MP)\in \U(\mbf{1}_{n\times m},\a,\b)$ is equivalent to $\P\in\U(\M,\a,\b)$. Therefore, Eq.  \eqref{eq:GWD_tensor_decompose} holds.
Since Eq. \eqref{eq:MGWD} can be rewritten as
\begin{align*}
         \mbf{L}_{mgw}(\M, \a,\b)= \min\limits_{\P\in \mbf{U}(\M,\a,\b)}\left< \MP,\hat{\C}\right>,
\end{align*}
this problem is equivalent to masked Wasserstein distance. Therefore, by proposition \ref{pro:MWD}, the solution is $$\P_{ij} = \u_i\M_{ij}\hat{\K}_{ij}\v_j.$$
Following the iterations methods proposed by \appcitet{peyre2016gromov}, with replacing the Sinkhorn iterations by log-domain Sinkhorn iterations, we summarize the computation of masked Gromov-Wasserstein distance in the Algorithm \ref{alg:MGWD}.

\begin{algorithm}[htb]
\def\u{\mathbf{u}}
\def\v{\mathbf{v}}
   \caption{Computing Masked Gromov-Wasserstein Distance}
   \label{alg:MGWD}
\begin{algorithmic}
   \STATE {\bfseries Input:} $\{x_i^s\}_{i=1}^{n}$, $\{x_i^t\}_{i=1}^{n}$, Masked Matrix $\M \in \{0,1\}^{n\times m}$, Marginals  $(\a,\b)$, Distance matrices $(\C,\bar{\C})$ 
  \STATE  Compute  $c_{\C,\bar{\C}}:= f_1(\C)\a\mbf{1}^{\top}_m + \mbf{1}_{n}\b^{\top}f_2(\bar{\C})^{\top} $ 
   \FOR{$i=1,2,3,...$}
   \STATE // compute the pseudo-cost matrix\\
   $\hat{\C}= c_{\C,\bar{\C}} - h_1(\C)(\MP) h_2(\bar{\C})^{\top}$
    \STATE Apply Algorithm \ref{alg:gtot} to solve transport plan $\P$.
    
   \ENDFOR
 
   \STATE  $\mathcal{D}_{mgw} = \left<\MP,\hat{\C} \right>$
   \STATE {\bfseries Output:} $\mathcal D_{mgw}$

\end{algorithmic}
\end{algorithm}

When we set $L(a,b)$ to be $(a-b)^2$,  $f_1(a)=a^2,f_2(b)=b^2,h_1(a)=2a,h_2(b)=b$. In fine-tuning, we can use $L(a,b)=(a-b)^2$ to calculate the Masked GWD.

\section{Missing Proofs.}
\subsection{The proof of Lemma \ref{le:uniform_stability}}
\restadeLemOne*
\paragraph{Proof:}
For any $(G,y)\in \mathcal{Z} (\mathcal{Z}:=\mathcal{X}\times \mathcal{Y})$, Let $h(G)$ denote the message passing phase of GNN $f(G)$ with graph $G$ as input ($h(G_i)_j$ denotes the $j$-th node representation of $G_i$ after message passing.) 
\begin{align*}
    &|l(f_S,z)-l(f_{S^i})| \\
    &= |\ell(f_{}(G),y)+\lambda\sum_{jk}P_{jk}(1-cos(h_{S}(G)_{j},h^{(s)}(G)_{k})/2  - \ell(f_{S^i}(G),y) -\lambda\sum_{jk}P_{jk}(1-cos(h_{S^i}(G)_{j},h^{(s)}(G)_{k})/2 | \\
    & \leq 2M + \lambda \left| tr(\left[\frac{1}{2}(1-\operatorname{RNorm}(h_S(G)) (\operatorname{RNorm}(h^{(s)}(G)))^{\top}) 
      -\frac{1}{2}(1-\operatorname{RNorm}(h_{S^i}(G)) (\operatorname{RNorm}(h^{(s)}(G)))^{\top})\right]^{\top}\P \right|,
\end{align*}
where $\operatorname{ RNorm(\cdot)}$ denotes the row normalization for calculating cosine. Let 
\begin{align*}
    \mathbf{C}_G := \frac{1}{2}(1-\operatorname{RNorm}(h_S(G)) (\operatorname{RNorm}(h^{(s)}(G)))^{\top}) 
      -\frac{1}{2}(1-\operatorname{RNorm}(h_{S^i}(G)) (\operatorname{RNorm}(h^{(s)}(G)))^{\top}),
\end{align*}
then $0\leq|\mathbf{C}_G|_{ij}\leq 1$. Thus

\begin{align*}
    |l(f_S,z)-l(f_{S^i})| & \leq
    2M + \lambda tr(|\mbf{C}_G|^{\top}\P)
    \leq 2M + \lambda \sqrt{tr(|\mbf{C}_G|^{\top}|\mbf{C}_G|)tr(\P^{\top}\P)} \\
    & = 2M + \lambda \sqrt{\sum_{ij}|\mathbf{C}_G|^2_{ij}\sum_{ij}P^2_{ij}} \leq 2M + \lambda \sqrt{\sum_{ij}1 (\frac{1}{|\V_G|^2}\sum_{ij}(|\V_G|P_{ij})^2)} \\
    & \leq 2M + \lambda \sqrt{\sum_{ij}(|\V_G|P_{ij})} \leq 2M + \lambda \sqrt{|\V_G|} \leq 2M + \lambda \sqrt{B}
\end{align*}

\subsection{The proof of Proposition \ref{pro:generalization_bound}
}

\begin{lemma}(\appcite{bousquet2002stability},Theorem 12)
\label{le:generalization_bound} 
Let $f$ be an algorithm with uniform stability $\beta$ with respect to a loss function $l$ such that $0 \leq l(f_S, z) \leq \mcal{Q}$, for all $z \in \mathcal{Z} $ and all sets $S$. Then, for any $m \ge 1$, and any $\delta \in (0,1)$, the bounds hold (separately) with probability at least $1-\delta$ over any random draw of the sample
$S$,
\begin{align}
    R(f_S)\leq R_m(f_S) + 2\beta + (4m\beta+\mcal{Q})\sqrt{\frac{\ln 1/\delta}{2m}}
\end{align}
\end{lemma}
\paragraph{Proof:} Substituting $\beta = 2M+\lambda\sqrt{B}$ into Lemma \ref{le:generalization_bound}, we can get the proposition \ref{pro:generalization_bound}.

\paragraph{Empirical Evidence.}
Empirically, Table \ref{tab:vertex_statistic} show the $B$ of MUV is the smallest among all the datasets. And our method gains a great improvement compared with fine-Tuning (Table  \ref{tab:molecular},\ref{tab:sup_GIN_molecular}. )
This corroborates  the correctness of our theorical results to some extent.

\newpage
\section{GTOT-MGWD Regularizer}\label{sec:MGWD_regularizer}

\begin{figure*}[ht]
    \centering
    \includegraphics[width=0.95\linewidth]{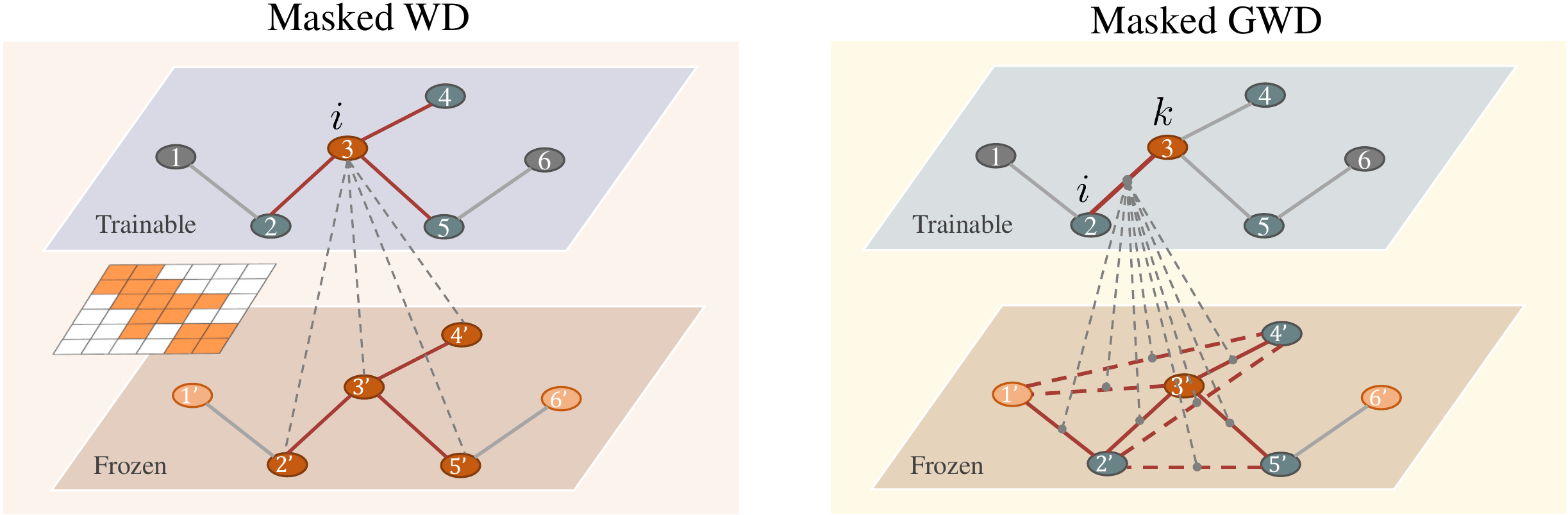}
     \caption{ An example of using masked WD and GWD on a graph. When one uses an adjacency matrix as a Masked matrix, the MWD can be rewritten as $\sum_{i}\sum_{j \in \mcal{N}(i)\bigcup \{i\}}\P_{ij}\C_{ij}$ ,where $i=3,\mcal{N}(i)=\{2',3',4',5'\}$,  and MGWD can be rewritten as   $\sum_{i}\sum_{j\in\mcal{N}(i)\bigcup \{i\}} \sum_{k}\sum_{l\in\mcal{N}(k)\bigcup \{k\}}L(\C_{ik},\bar{\C}_{jl}) \P_{ij}\P_{kl}$ , where $i=2,\mcal{N}(i)\bigcup \{i\}=\{1',2',3'\}$ and $k=3,\mcal{N}(k)\bigcup \{k\}=\{2',3',4',5'\}$. }
    \label{fig:maskedWD_gwd_details}
        \vspace{-2mm}
\end{figure*}

Along with the design of the GTOT-Regularizer, we propose the GTOT-MGWD Regularizer with adjacency matrix as mask matrix in Eq. \eqref{eq:MGWD}. 
Unlike the GTOT-Regularzer concentrating on preserving the local node invariances, GTOT-MGWD Regularizer expects to preserve the local edge invariances (Fig. ~\ref{fig:maskedWD_gwd_details}).

\subsection{Objective Function}
Following the usage of GTOT Regularizer, the GTOT-MGWD Regularizer can be used as follows
\begin{align}
    \small
    \sum_{i=1}^{m} (\ell(f(G_i),y_i) + \beta \mbf{L}_{mgw}(\A_i,\q,\q)
    )
\end{align}

\begin{figure*}[ht]
    \centering
    \includegraphics[width=0.99\linewidth]{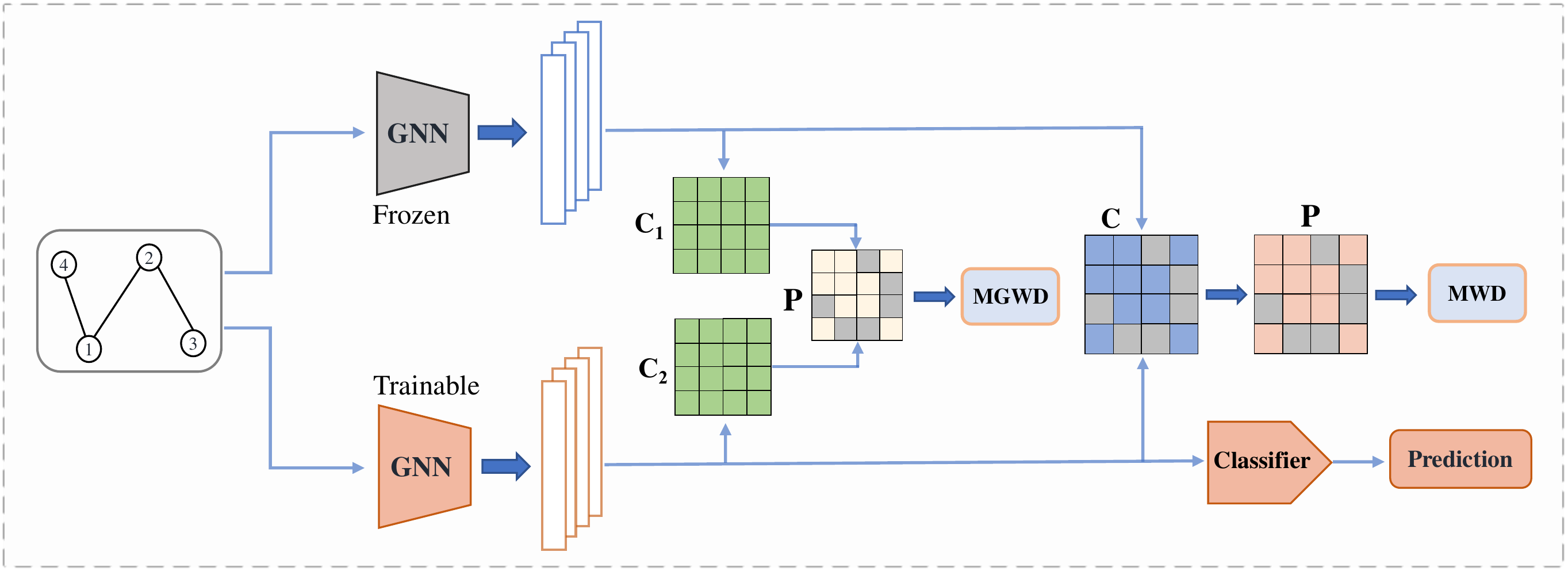}
     \caption{ The finetuning framework after combining two regularizers. The gray lattice of $\P$ indicates that
     $\P_{ij}=0$ when the vertex pair $(v_i,v_j)$ is not adjacent. Assume that the input graph has self-loops.  }
    
    \label{fig:gtot_framework_all}
        \vspace{-2mm}
\end{figure*}

 Intuitively, one can jointly use MWD and MGWD to achieve better fine-tuning performance via the simultaneous alignment of node and edge information. Here, we provide a naive method to combine the two regularizers (Figure \ref{fig:gtot_framework_all}) named GTOT-(MWD+MGWD)
\begin{align}
    \small
    \sum_{i=1}^{m} (\ell(f(G_i),y_i)+ 
    \lambda \mbf{L}_{mw}(\A_i,\q,\q)+ \beta \mbf{L}_{mgw}(\A_i,\q,\q)
    )
\end{align}

\subsection{Experiments}
\paragraph{Comparison of Different Fine-Tuning Strategies}
Table \ref{tab:molecular_gwd_all} shows the results of GTOT-MGWD and GTOT-(MWD+MGWD). As we can see in the table, GTOT-MGWD slightly outperforms GTOT-MWD on average, indicating that GTOT-MGWD is also an efficient regularizer for finetuning GNNs. However, GTOT (MWD+MGWD) cannot obtain better performance than either of the two regularizers (GTOT-MWD, GTOT-(MWD+MGWD)). More effective joint methods are worth exploring in the future.

\paragraph{Ablation studies.} We do the ablation study for MGWD to investigate the impact of the adjacency matrix and the results show in table \ref{tab:ab_gwd_gin}. It is clear that the introduced Mask matrix $\A$ (adjcancy matrix) of GTOT-MGWD is effective.

\begin{table*}[th]
\def\p{$\pm$} 
\setlength\tabcolsep{3pt}
\centering
\vspace{-2mm}
\caption{Test ROC-AUC (\%) of GIN (contexpred) on downstream molecular property prediction benchmarks.}
\vspace{-0mm}
\scalebox{0.74}{
\begin{tabular}{l|cccccccc|c}
\toprule 
Methods & BBBP & Tox21 & Toxcast& SIDER & ClinTox & MUV & HIV & BACE & \textbf{Average}\\
\midrule
Fine-Tuning~(baseline) & 68.0$\pm$2.0 & \underline{75.7\p 0.7} & 63.9\p 0.6 & 60.9\p 0.6 & 65.9\p 3.8 & 75.8\p 1.7 & 77.3\p 1.0 & 79.6\p 1.2 & 70.85 \\
\midrule
L2\_SP~\cite{xuhong2018explicit} & 68.2$\pm$0.7 & 73.6$\pm$0.8 & 62.4$\pm$0.3 & 61.1$\pm$0.7 & 68.1$\pm$3.7 & 76.7$\pm$0.9 & {75.7$\pm$1.5} & 82.2$\pm$2.4 & 70.25 \\
DELTA\cite{li2018delta} & 67.8$\pm$0.8 & 75.2$\pm$0.5  & 63.3$\pm$0.5 & 62.2$\pm$0.4 & \textbf{73.4$\pm$3.0} & \textbf{80.2$\pm$1.1} & 77.5$\pm$0.9 & 81.8$\pm$1.1 & 72.68 \\
Feature(DELTA w/o ATT) & 61.4$\pm$0.8 & 71.1$\pm$0.1 & 61.5$\pm$0.2 & \underline{62.4$\pm$0.3} & 64.0$\pm$3.4 & {78.4$\pm$1.1} & 74.0$\pm$0.5 & 76.3$\pm$1.1 & 68.64 \\
BSS\cite{chen2019catastrophic} & 68.1$\pm$1.4 & \textbf{75.9$\pm$0.8} &  \underline{63.9$\pm$0.4} & 60.9$\pm$0.8 & {70.9$\pm$5.1} & 78.0$\pm$2.0 & \underline{77.6$\pm$0.8} & \underline{82.4$\pm$1.8} & 72.21 \\
StochNorm~\cite{kou2020stochastic} & \underline{69.3$\pm$1.6} & 74.9$\pm$0.6 & 63.4$\pm$0.5 & 61.0$\pm$1.1 & 65.5$\pm$4.2 & 76.0$\pm$1.6 & \underline{77.6$\pm$0.8} & 80.5$\pm$2.7 & 71.03\\
\midrule
GTOT-MWD~(Ours) & \textbf{70.0$\pm$2.3}$\uparrow_{2.0} $ & {75.6$\pm$0.7}$\downarrow_{0.1} $ & \textbf{64.0$\pm$0.3}$\uparrow_{0.1}$ & \textbf{63.5$\pm$0.6}$\uparrow_{2.6}$ & \underline{72.0$\pm$5.4}$\uparrow_{6.1}$ & \underline{80.0$\pm$1.8}$\uparrow_{4.2}$ & \textbf{78.2$\pm$0.7}$\uparrow_{0.9}$ & \textbf{83.4$\pm$1.9}$\uparrow_{3.8}$ &  \textbf{73.34}$\uparrow_{2.49}$ \\

\midrule
GTOT-MGWD~(Ours) & \textbf{69.8$\pm$2.4}$\uparrow_{1.8}$ & 75.4$\pm$0.3$\downarrow_{0.3}$ & 64.2$\pm$0.4$\uparrow_{0.3}$ & 61.9$\pm$1.4$\uparrow_{1.0}$ & \textbf{75.3$\pm$6.0}$\uparrow_{9.4}$ & 79.3$\pm$1.4$\uparrow_{3.5}$ & \textbf{78.3$\pm$0.8}$\uparrow_{1.0}$ & 83.2$\pm$1.7$\uparrow_{3.6}$ & 73.43 \\
GTOT(MWD+MGWD)(Ours) & 69.5$\pm$2.3 & 75.2$\pm$0.4 &  64.3$\pm$0.5 & 62.4$\pm$1.0 & 72.8$\pm$3.6 & 80.2$\pm$1.4 & 78.0$\pm$0.7 & 84.1$\pm$0.6 & 73.31 \\ 
\bottomrule
    \end{tabular}
    }
    \vspace{-0mm}
    \label{tab:molecular_gwd_all}    
\end{table*}

\begin{table}[ht]
\def\p{$\pm$} 
\centering
\vspace{-2mm}
\caption{Test ROC-AUC (\%) of GIN (contextpred) on downstream molecular property prediction benchmarks.(BPTs means Binary prediction tasks.)}
\vspace{-0mm}
\scalebox{0.85}{
\begin{tabular}{c|cccc}
\toprule 
Methods & BBBP & Tox21 & Toxcast& SIDER   \\
\# BPTs & 1 & 12 & 617 & 27 \\
\midrule
GTOT-MGWD($\mbf{1}_{n\xx n}$)  & 68.6$\pm$1.9 & \textbf{75.3$\pm$0.5} & 63.6$\pm$0.5  & 61.4$\pm$1.0 \\
 GTOT-MGWD($\A$) & \textbf{69.1$\pm$2.5} & {75.2$\pm$0.8} & \textbf{64.2$\pm$0.4} & \textbf{62.0$\pm$1.2} \\
\midrule
 Methods & ClinTox & MUV & HIV & BACE \\
 \# BPTs & 2 & 17 & 1 & 1 \\
 \midrule
 GTOT-MGWD($\mbf{1}_{n\xx n}$) & 62.8$\pm$9.5  & 75.8$\pm$2.0 & \textbf{77.8$\pm$0.9} & 83.2$\pm$1.7 \\
 GTOT-MGWD($\A$) & \textbf{68.5$\pm$7.2} & \textbf{77.2$\pm$1.8} & {77.7$\pm$0.9} & \textbf{83.4$\pm$2.2} \\ 
\bottomrule
    \end{tabular}
    }
    \vspace{-0mm}
    \label{tab:ab_gwd_gin}    
\end{table}

\newpage

\section{Details of Experiments.}

For the cost matrix $\C_{ij}=\frac{1}{2}(1-cosine(\x_i^S,\x_j^T))$ in GTOT regularizer, we use the maximum normalization $2\C/\C_{max}$ instead to scale the matrix elements to $[0,1]$. Noted that this operation has essentially no effect on the results. 

 In our experiments, we use the output from the last linear layer ('gnn.gnns.4.mlp.2') of GNN message passing to calculate the GTOT distance as well as the baselines Feature(Delta w/o  ATT) and DELTA.
 Indeed, the GTOT regularizer can be used in any middle layer of the backbone.
 
 We follow \appcite{hu2020strategies}'s data split (Train:validation:test = 8:1:1) to fine-tuning pretrained models for downstream tasks.
Each method we run $10$ random seeds~$(0-9)$ and we report the mean $\pm$ standard deviation.
\subsection{Hyper-parameter Strategy}
\label{subsec:hyper-parameter}
We use cross-entropy loss and Adam optimizer with early stopping with patience of 20 epochs to train GTOT-Tuning models.  For other hyper-parameters, 
We use grid search strategies and the range of hyper-parameters listed in Table \ref{tab:hyper_parameter}.
 
\begin{table}[htbp] 
\centering
\caption{Hyper-parameter search range for GTOT-Tuning. }
\vspace{-1mm}
\resizebox{0.8 \linewidth}{!}{
\begin{tabular}{c|l|c}
\toprule
      & Hyper-parameter        & Range               \\ 
\midrule
    & $\lambda$ (GTOT regularizer) & \{1e-7,1e-6,5e-6,1e-5,5e-5,1e-4,5e-4,1e-3,5e-3,1e-2,5e-2,1e-1,5e-1,1,5,10\} \\
    & $\beta$ (MGWD regularizer Sec. \ref{sec:MGWD_regularizer})    & \{1e-7,1e-6,5e-6,1e-5,5e-5,1e-4,5e-4,1e-3,5e-3,1e-2,5e-2,1e-1,5e-1,1,5,10\} \\
    & Learning rate        & \{0.001, 0.005, 0.01\}                  \\\
    & Weight decay & \{1e-2, 1e-3,1e-4,5e-4, 1e-5,1e-6,1e-7,0\}  \\
    & Dropout rate        & \{0,0.05,0.1,0.15,0.2,0.25,0.3,0.35, 0.4,0.45, 0.5\} \\
    & Batch size & \{16,32,64,128\}  \\
    \midrule
    & Optimizer & Adam  \\
    & Epoch & 100 \\
    & Early stopping patience & 20 \\
    & GPU & 
Tesla V100 \\
  
\bottomrule                     
\end{tabular}                           
}
\label{tab:hyper_parameter}
\end{table}

\subsection{Datasets} The downstream dataset statistics are summarized in Table \ref{tab:vertex_statistic}.
 
 \begin{table}[ht]
\def\p{$\pm$} 
\centering
\vspace{-2mm}
\caption{Statistical information on downstream task datasets. We provide (median, max, min, mean, std) number of graph vertices in different datasets.}
\vspace{-0mm}
\scalebox{0.95}{
\begin{tabular}{l|rr|rrrrr}
\toprule 
dataset &\# Molecules & \# training set & median & max & min & mean & std \\
\midrule
 BBBP & 2039 &1631 & 22 & 63 & 2 & 22.5 & 8.1 \\
 Tox21 & 7831& 6264 & 14 & 114 & 1 & 16.5 & 9.5 \\
Toxcast & 8575 & 6860 & 14 & 103 & 2 & 16.7 & 9.7 \\
SIDER & 1427& 1141 & 23 & 483 & 1 & 30.0 & 39.7 \\
ClinTox & 1478 & 1181 & 23 & 121 & 1 & 25.5 & 15.3 \\
MUV & 93087 & 74469 & 24 & 44 & 6 & 24.0 & 5.0 \\
HIV & 41127 & 32901 & 23  & 222 & 2 & 25.3 & 12.0 \\
BACE & 1513 & 1210 & 32 & 66 & 10 & 33.6 & 7.8 \\
\bottomrule
    \end{tabular}
    }
    \vspace{-0mm}
    \label{tab:vertex_statistic}    
\end{table}

\subsection{Backbones}

\paragraph{Graph Neural Networks}
\def\e{\mbf{e}}
Generalizing the convolution operator to irregular domains is typically expressed as a neighborhood aggregation or message passing scheme~\appcite{yadati2020neural}. With $\x_i^{(k-1)}\in \R^{F}$  denoting node features of node $i$ in layer $(k-1)$  and $\e_{j,i}\in \R^{D}$ denoting (optional) edge features from node $j$ to node $i$, message passing graph neural networks (MPNNs) can be described as

\begin{align}
\label{eq:MPNN}
    \mathbf{x}_{i}^{(k)}=\gamma^{(k)}\left(\mathbf{x}_{i}^{(k-1)}, \square_{j \in \mathcal{N}(i)} \phi^{(k)}\left(\mathbf{x}_{i}^{(k-1)}, \mathbf{x}_{j}^{(k-1)}, \mathbf{e}_{j, i}\right)\right)
\end{align}
where $\square$ denotes a differentiable, permutation invariant function, e.g., sum, mean or max, and $\phi$ and $\gamma$ denote differentiable functions such as MLPs (Multi Layer Perceptrons).

Our GTOT-Regularization acts on the output node embeddings of message passing $\{\x_1^{(k)},...,\x_{|\V|}^{(k)}\}$. GCN~\appcite{kipf2016semi} and GIN~\appcite{xu2018powerful} can be written as MPNNs, one can refer to PyG \footnote{\url{https://pytorch-geometric.readthedocs.io/en/latest/notes/create_gnn.html}}.

\paragraph{GIN~(contextpred) and GIN~(supervised\_contextpred).}
The GIN~(contextpred) and GIN~(supervised\_contextpred) we directly adopt the released models from \appcitet{hu2020strategies}. For node-level self-supervised pre-training, the pre-trained data are 2 million unlabeled molecules sampled from the ZINC15 database \appcite{sterling2015zinc}.  For graph-level multi-task supervised pre-training, the pre-trained dataset is a pre-processed ChEMBL dataset~\appcite{mayr2018large}, containing 456K molecules with 1310 kinds of diverse and extensive biochemical assays.

The GCN~(contextpred) used in additional experiments is also from \appcitet{hu2020strategies}.

The details of these architectures can be found in Appendix A of \appcitet{hu2020strategies}.

\subsection{Baselines}

 \paragraph{Fine-Tuning~(Baseline)}
 We reuse the AUC reported by \appcite{hu2020strategies}.
\\
\\
We refer to the transfer learning package\footnote{\url{https://github.com/thuml/Transfer-Learning-Library}} for the implementations of the following baselines.

\paragraph{L2\_SP}\appcite{xuhong2018explicit}
The L2\_SP is a weights regularization. The key concept of L2\_SP penalty is "starting point as reference":
\begin{align}
   \Omega(W) =   \frac{\beta}{2}\|W^S-W^T\|_2^2 +\frac{\beta}{2} \|W_{\bar{T}}\|_2^2
\end{align}
where $W^S$ is the pre-trained weight parameters of the shared architecture, $W^T$ is weight parameters of the model, $W_{\bar{T}}$ is weight parameters of the task-specific classifier, $\beta$ is a trade-off hyper-parameter to control the strength of the penalty. L2\_SP penalty tries to drive weight parameters to pre-trained values.

\paragraph{DELTA}\appcite{li2018delta}
Based on the key insight of "unactivated channel re-usage", Delta is proposed as a regularized transfer learning framework. Specifically, DELTA selects the discriminative features from higher layer outputs with a supervised attention mechanism. 
However, it is designed for Convolutional Neural Networks(CNN). Therefore, we have to make some adjustments to accommodate GNNs.
The channels of CNN correspond the column of the learnable parameter matrix $\Theta \in \R^{d_l\xx\ d_{l+1}}$ in GIN or GCN~(i.e. the $\gamma$ in Eq. \eqref{eq:MPNN}), where $\Theta$ totally has $d_{l+1}$ filters and would generates $d_{l+1}$ feature maps. Each feature map is a $n$-dim vector, so here we call it Feature Vector~(FV). Therefore, the Delta is calculated as follows for adapting to GNNs.
\begin{align*}
    \Omega(W) = \gamma \Omega'(W^T,W^S, \x_i,y_i,z) + \kappa\Omega''(W_{\bar{T}})
\end{align*}
where 
\begin{align*}
    \Omega'(W^T,W^S, \x_i,y_i,z) :=  \sum_{j}^{d_{l+1}} D_j(z,W^S,\x_i,y_i)\|\text{FV}_j(z,W^T,\x_i)-\text{FV}_j(z,W^S,\x_i)\|_2^2
\end{align*}
and 
\begin{align}
\label{eq:D_j}
D_j(z,W^S,\x_i,y_i):=\text{softmax}(\mcal{L}(z(\x_i,W^{S/j} ), y_i)- \mcal{L}(z(\x_i,W^{S} ), y_i))    
\end{align}
where $\mcal{L}$ is the cross-entropy loss in our experiments.
$z$ is the model, $\Omega'$ is behavioral regularizer, $\Omega''$ constrains the $L_2$-norm of the private parameters in $W^T$; $D_j(z,W^S,\x_i,y_i)$ refers to the behavioral difference between the two feature vectors and
the weight assigned to the $j$-th filter and the $i$-th graph (for $1\leq j\leq d_{l+1}$); $\gamma$ and $\kappa$ are trade-off hyper-parameters to control the strength of the two regularization terms.

\paragraph{Feature(DELTA w/o ATT)}
When the attention coefficients $D_j(z,W^S,\x_i,y_i)$ (Eq. \eqref{eq:D_j} ) in DELTA is $1/2$, DELTA will become the \textit{Feature(DELTA w/o ATT)} regularization. Actually, it is equivalent to  a node-to-node representation regularization
\begin{align}
    \Omega'(W^T,W^S, \x_i,y_i,z) :=  \frac{1}{2}\sum_{j}^{n} \|\text{NFV}_j(z,W^T,\x_i)-\text{NFV}_j(z,W^S,\x_i)\|_2^2
\end{align}
where $\text{NFV}_j(z,W^T,\x_i)\in\R^{1\xx d_{l+1}}$ denote the $j$-th Node Feature Vector~(NFV) produced by message passing neural networks $z_{W^T}(\x_i)$.
\begin{remark}
Following the notations from DELTA and Feature(DELTA w/o ATT), our GTOT Regularizer can be rewritten as 
\begin{align}
    \Omega'(W^T,W^S, \x_i,y_i,z) := \min\limits_{\P\in \mbf{U}(\A,\q,\q)} \frac{1}{2}\sum_i^n\sum_{j}^{n}\A_{ij} \P_{ij} (1- \operatorname{cosine}(\operatorname{NFV}_i(z,W^T,\x_i), \operatorname{NFV}_j(z,W^S,\x_i)))
\end{align}
\textbf{This suggests that our method focuses on the distance of node-level embeddings, rather than the distance of channel-level features. }
\end{remark}

\paragraph{BSS.} Inheriting  from \appcitet{chen2019catastrophic}, we use the SVD to compute the feature matrix $\mbf F=[\mbf{g}_1,\mbf{g}_2,...,\mbf{g}_b]$ from the output of the readout layer in GNNs, where $\mathbf g_i$ denotes the $i$-th graph representation in a batch. Finally, the BSS regularization is :
\begin{align*}
    \mcal{L}_{bss} = \eta \sum_{i=1}^{k}\sigma_{-i}^2
\end{align*}
where $\eta$ is a trade-off hyper-parameter to control the strength of spectral shrinkage, $k$ is the number of
singular values to be penalized, and $\sigma_{-i}$ refers to the $i$-th smallest singular value of $\mbf F$.

\paragraph{Stochnorm.} The Stochnorm is used directly from \appcitet{kou2020stochastic}  without any modification.

\subsection{Additional Experiments}\label{subsec:additional_experiments}
\begin{figure*}[htbp]
\centering
    \subfigure{
    \begin{minipage}[s]{0.4\linewidth}
    \centering
    \includegraphics[width=1\linewidth]{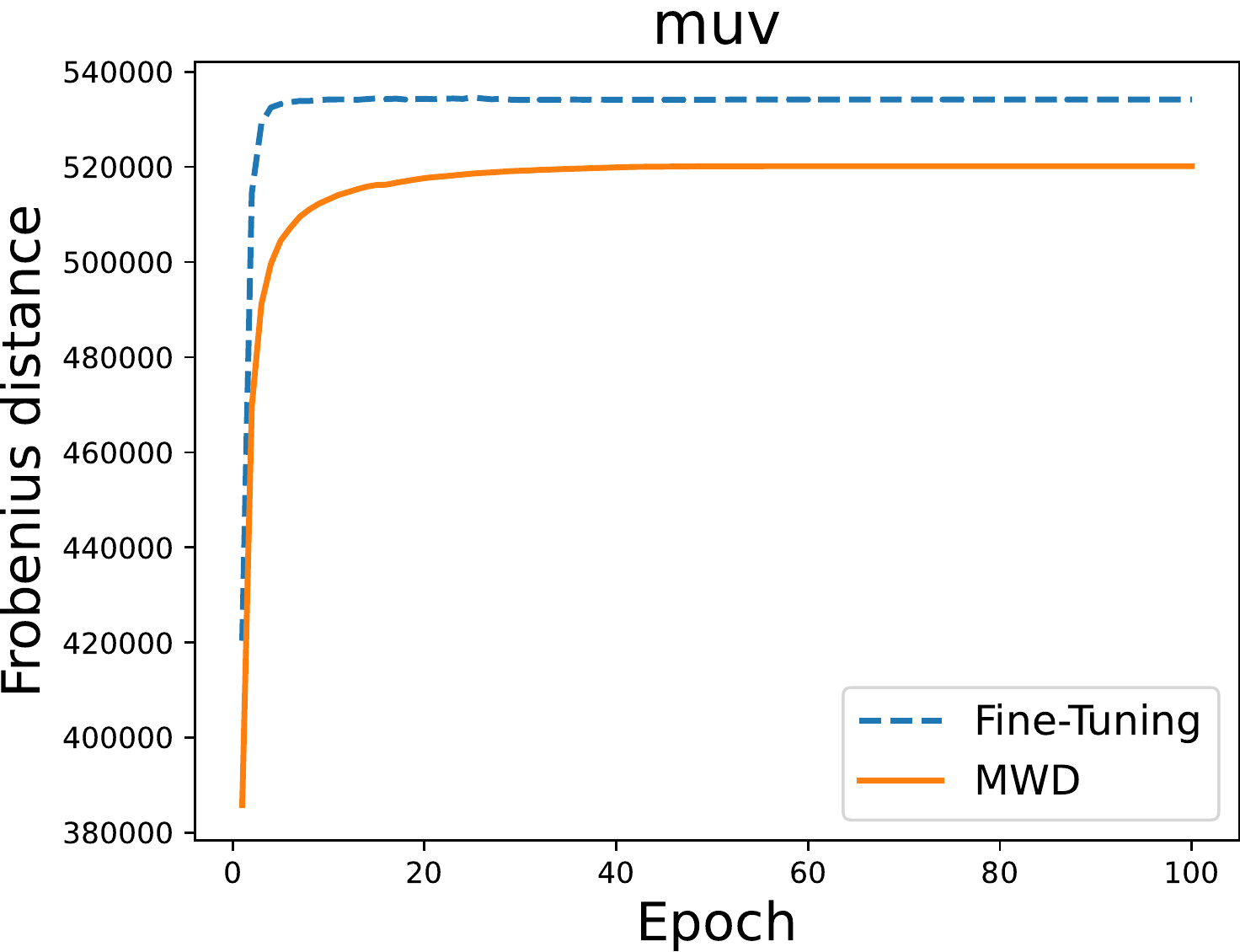}
    \end{minipage}%
    }%
    \quad
    \subfigure{
    \begin{minipage}[s]{0.4\linewidth}
    \centering
    \includegraphics[width=1\linewidth]{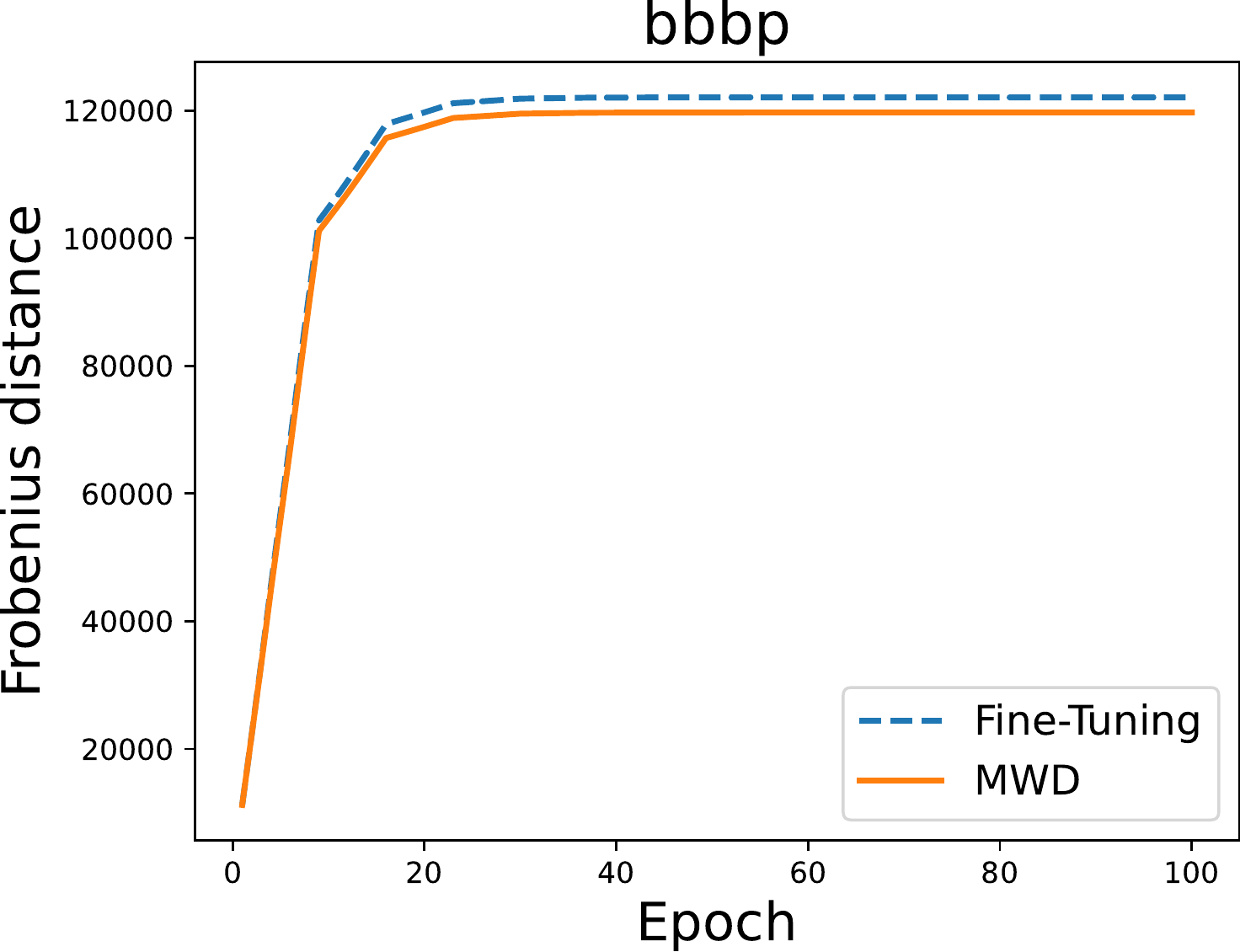}
    \end{minipage}%
    }%
    \quad
    \subfigure{
    \begin{minipage}[s]{0.4\linewidth}
    \centering
    \includegraphics[width=1\linewidth]{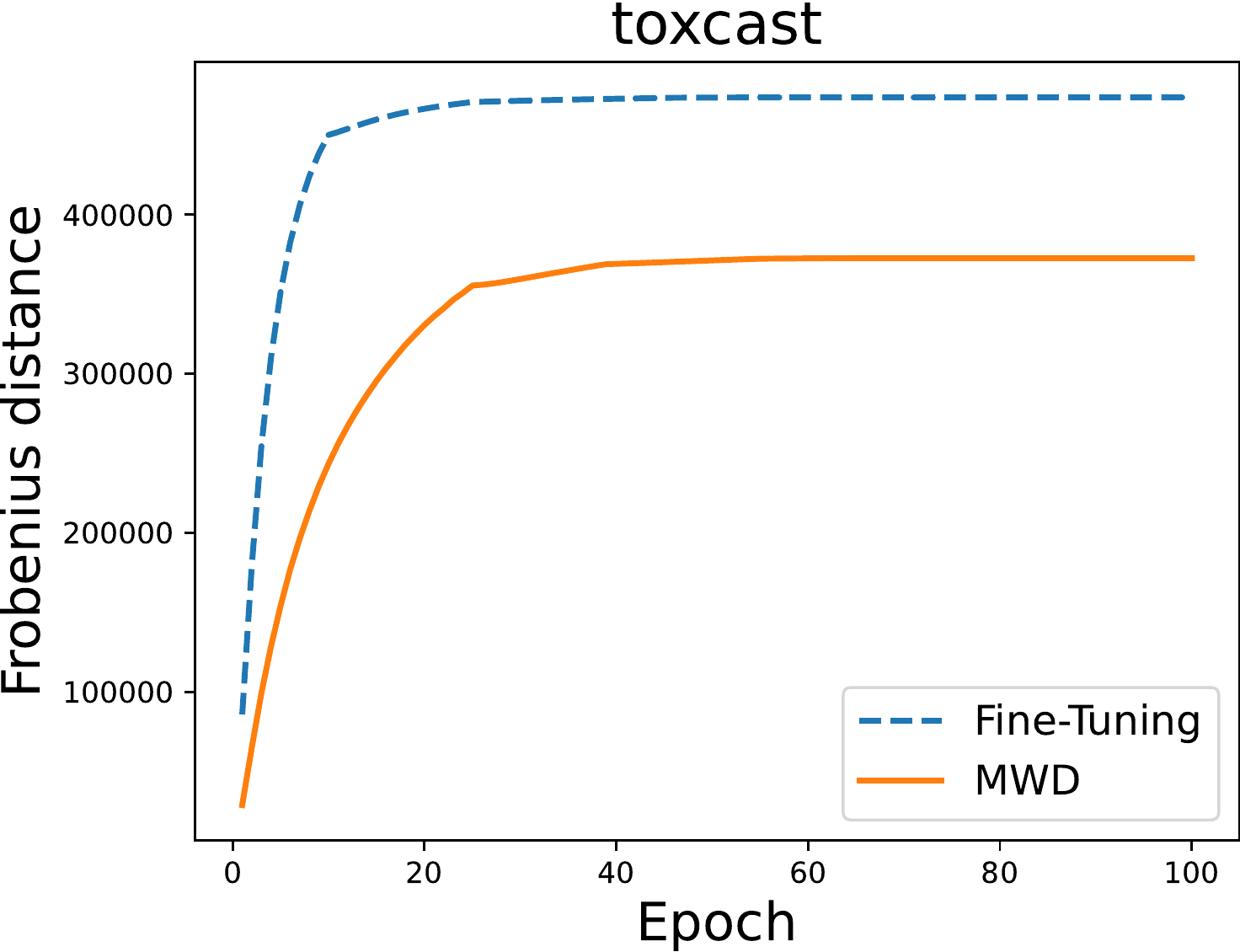}
    \end{minipage}
    }%
    \subfigure{
    \begin{minipage}[s]{0.4\linewidth}
    \centering
    \includegraphics[width=1\linewidth]{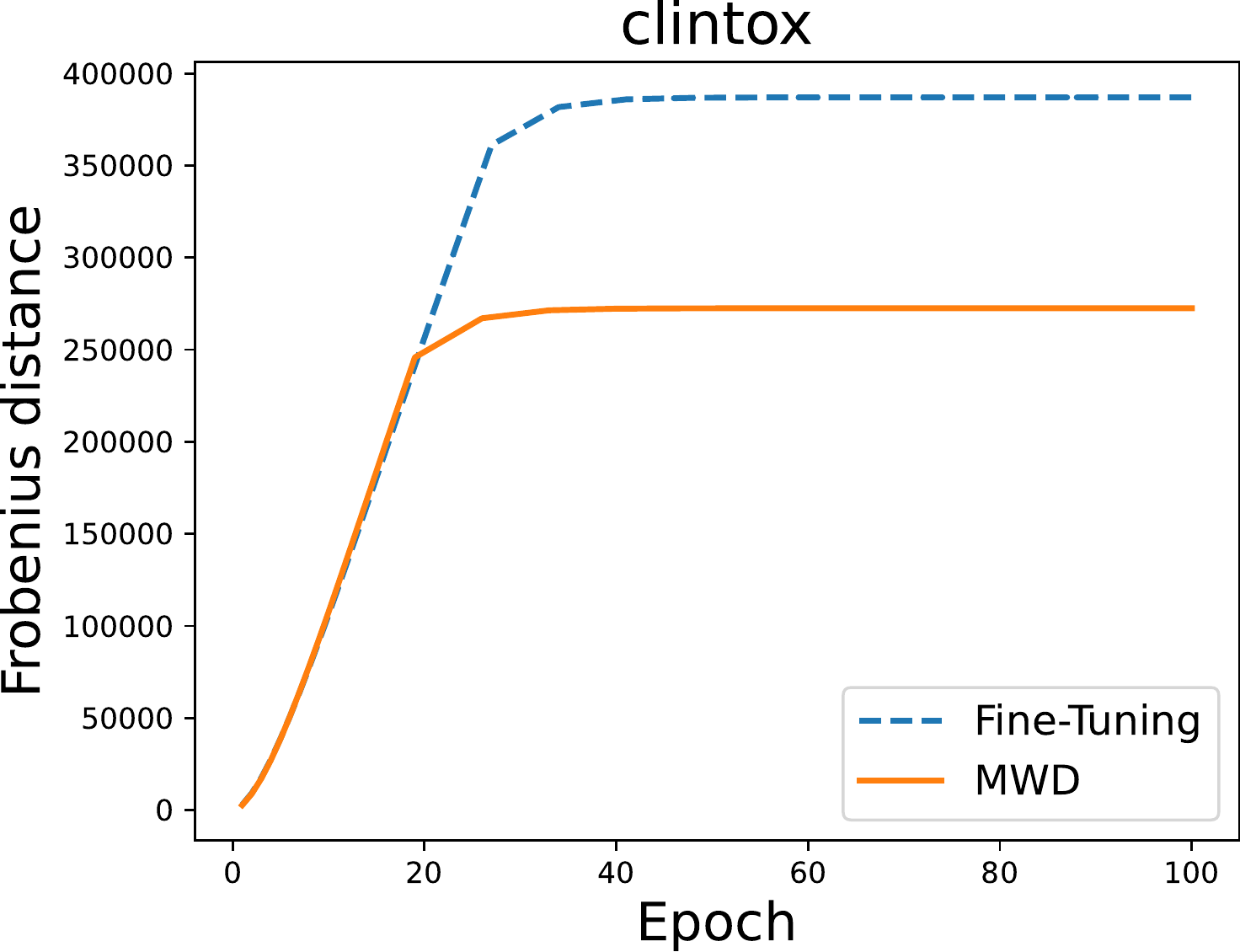}
    \end{minipage}
    }

\caption{Weights Distance curves between pre-trained initialization weights and fine-tuned weights. 
Comparing the vanilla finetuning, our GTOT regularizer can implicitly constrain the weights to deviate from the pre-trained parameters, alleviating the catastrophic forgetting to some extent. }
\label{fig:additional_catastropic_forgetting}
\end{figure*}

\paragraph{Catastrophic forgetting.} We conduct experiments to verify that our GTOT-Tuning is able to implicitly constrain the distance between fine-tuned and pre-trained weights, alleviating the Catastrophic forgetting issue in fine-tuning. The results are shown in Fig. \ref{fig:additional_catastropic_forgetting}.

\begin{figure}[htbp]
\centering
    \subfigure{
    \begin{minipage}[s]{0.4\linewidth}
    \centering
    \includegraphics[width=1\linewidth]{images/BACE_seed42_GTOT.pdf}
   \caption*{(a)}
    \end{minipage}%
    }%
    \quad
    \subfigure{
    \begin{minipage}[s]{0.4\linewidth}
    \centering
    \includegraphics[width=1\linewidth]{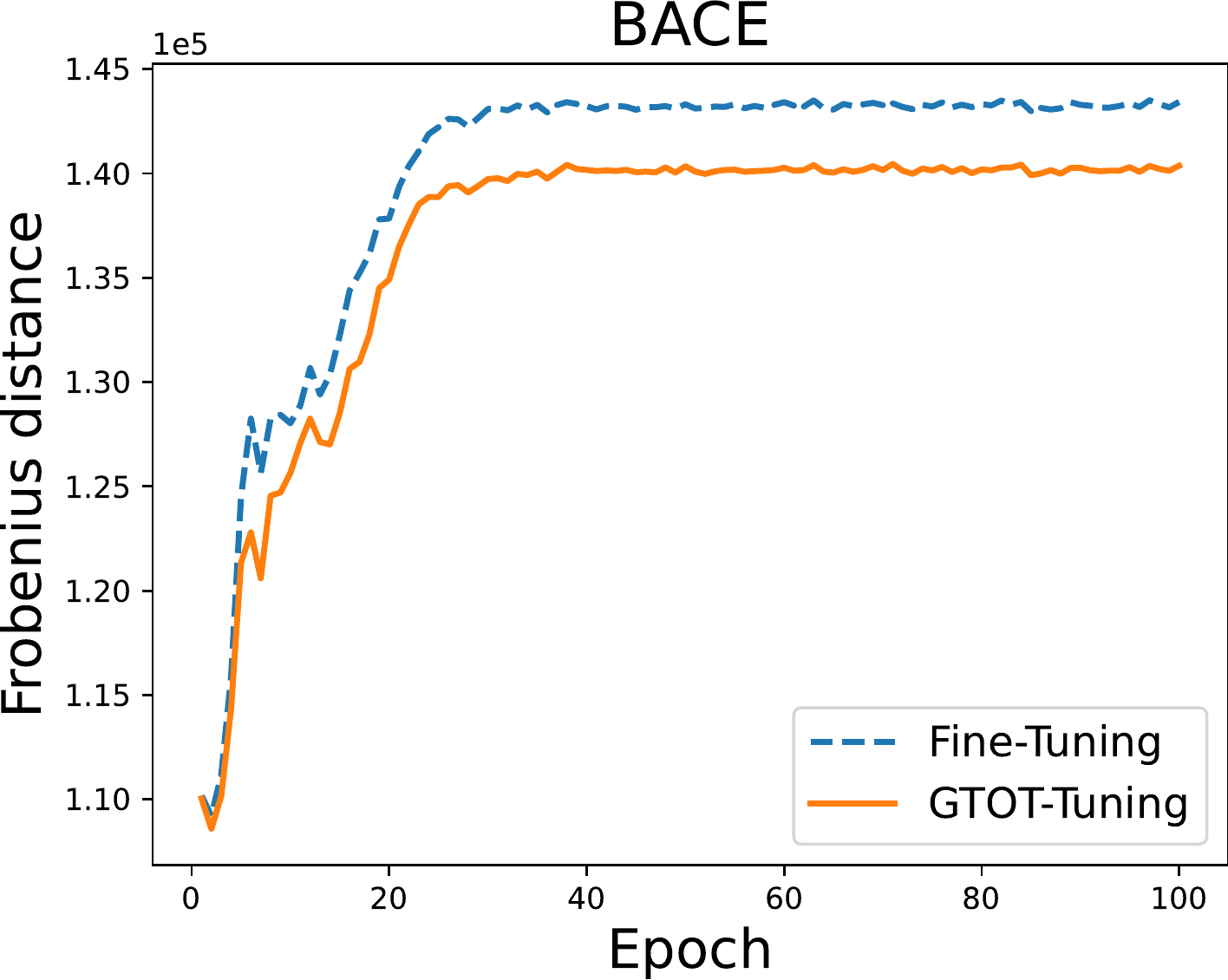}
   \caption*{(b)}
    \end{minipage}%
    }%
    \\
    \subfigure{
    \begin{minipage}[s]{0.4\linewidth}
    \centering
    \includegraphics[width=1\linewidth]{images/MUV_seed0_GTOT.pdf}
   \caption*{(c)}
    \end{minipage}
    }%
    \quad
    \subfigure{
    \begin{minipage}[s]{0.4\linewidth}
    \centering
    \includegraphics[width=1\linewidth]{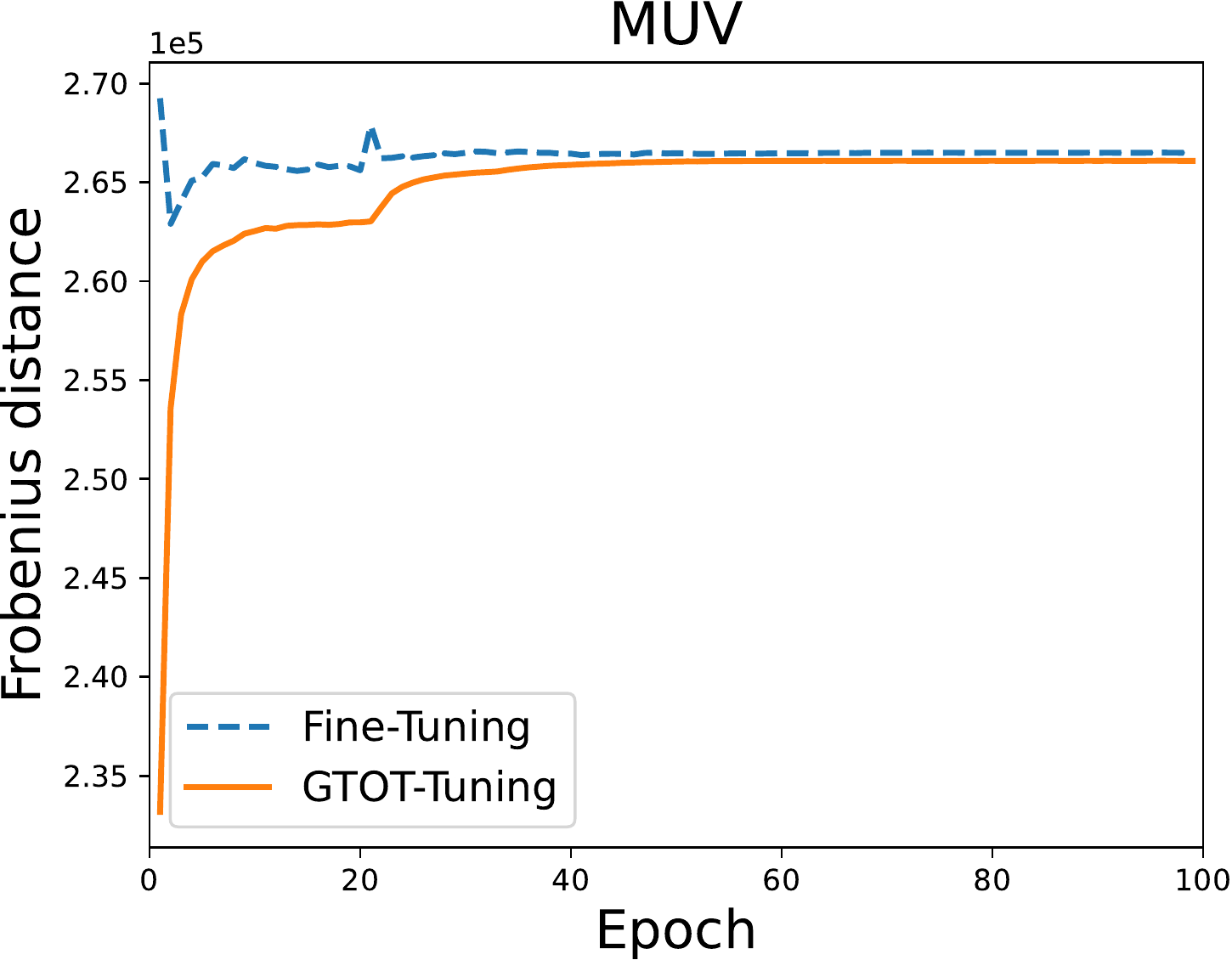}
  \caption*{(d)}
    \end{minipage}
    }
\vspace{-3mm}
\caption{(\textbf{left} (a)(c)): Weights distance between pre-trained initialization weights and fine-tuned weights. (\textbf{right} (b)(d)): Representations distance between vertex features extracted from pre-trained model and fine-tuned model. Benefitting from the soft alignment of MOT, the GTOT regularizer constrains the behaviors but allows the weights to be self-adapted to the downstream task.  }
\vspace{-2mm}
\label{fig:adaptive_domain_gap_additional}
\end{figure}

\paragraph{Domain Adaptation.}
As alluded to in the main text, GTOT is capable of adaptively constraining the distance of weights according to the domain gap~(Table~\ref{fig:adaptive_domain_gap_additional}), which also make it an ideal regularizer to handle the out of distribution problems~(when the domain gap is large).

\begin{figure*}[htbp]
\centering
    \subfigure{
    \begin{minipage}[s]{0.48\linewidth}
    \centering
    \includegraphics[width=1\linewidth]{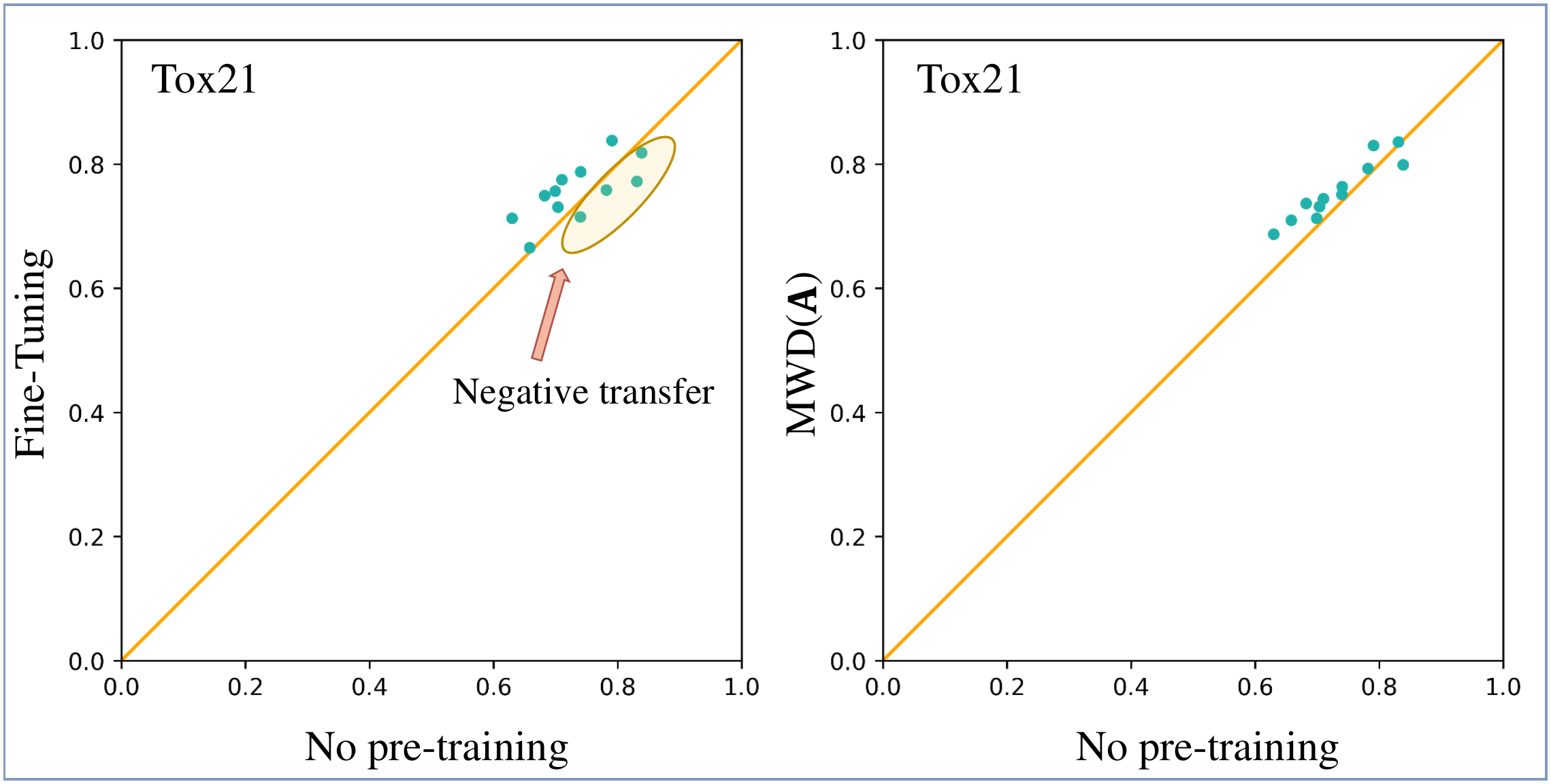}
    \end{minipage}%
    }%
    \quad
    \subfigure{
    \begin{minipage}[s]{0.48\linewidth}
    \centering
    \includegraphics[width=1\linewidth]{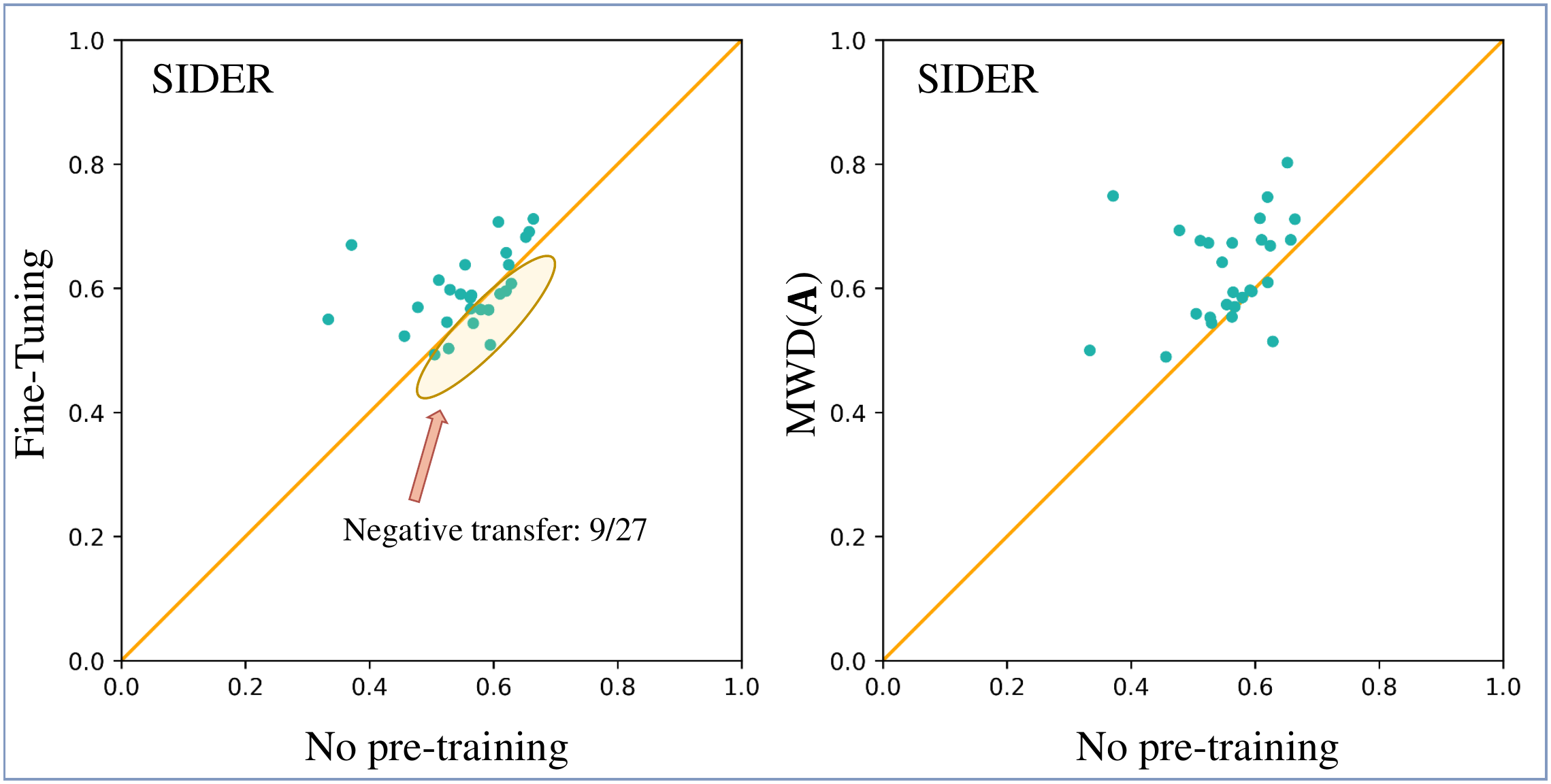}
    \end{minipage}%
    }%
   
\centering

\caption{Scatter plot
comparisons of ROC-AUC scores for a pair of fine-tuning strategies on the multi-task datasets (Tox21,SIDER). Each point represents a particular downstream task. There are many downstream tasks where vanilla fine-tuning performs worse than non-pre-trained model, indicating negative transfer. Meanwhile, negative transfer is alleviated when the GTOT regularization is used. }
\label{fig:additional_negative_transfer}
\end{figure*}

\paragraph{Negative Transfer.}
We provide additional results on multi-task datasets~(Fig. \ref{fig:additional_negative_transfer}) to support our conclusion in section ~\ref{sec:why_work} of main text.

\paragraph{}

\paragraph{GCN(contextpred).} 
Table \ref{tab:molecular_gcn} shows the results of GCN~(contextpred). It is clear that our methods have performance gains compared to the baseline in most cases.
\begin{table*}[th]
\def\p{$\pm$} 
\setlength\tabcolsep{5pt} 
\centering
\vspace{-2mm}
\caption{Test ROC-AUC (\%) of GCN (contexpred) on downstream molecular property prediction benchmarks.}
\vspace{-0mm}
\scalebox{0.84}{
\begin{tabular}{l|cccccccc|c}
\toprule 
Methods & BBBP & Tox21 & Toxcast& SIDER & ClinTox & MUV & HIV & BACE & \textbf{Average}\\
\midrule
Fine-Tuning~(baseline) & 68.5$\pm$1.3	& 73.8$\pm$0.7	& 63.7$\pm$0.9	& 60.3$\pm$1.1 &	68.1$\pm$6.9	& 75.3$\pm$1.3 & 	77.4$\pm$1.1 & 79.5$\pm$4.3 & 70.83 \\
GTOT-Tuning~(Ours) & 70.3$\pm$1.7  &   75.5$\pm$0.5 & 64.2$\pm$0.5 &  60.2$\pm$1.0 & 78.6$\pm$2.8 & 79.3$\pm$1.6 & 77.8$\pm$0.9  & 81.5$\pm$1.7 & 73.44 \\
GTOT-MGWD~(Ours) & 69.9$\pm$1.9 & 74.0$\pm$0.7 & 64.0$\pm$0.5 & 60.2$\pm$0.9 & 76.2$\pm$5.2  &80.8$\pm$1.1  & 79.1$\pm$0.9 & 81.8$\pm$1.3 & 73.25  \\
\bottomrule
    \end{tabular}
    }
    \vspace{-0mm}
    \label{tab:molecular_gcn}    
\end{table*}

\newpage
\subsection{Hyper-parameter Sensitivity}\label{subsec:Hyper-parameter_ Sensitivity}

\begin{figure*}[htbp]
\centering
    \subfigure{
    \begin{minipage}[s]{0.5\linewidth}
    \centering
    \includegraphics[width=1\linewidth]{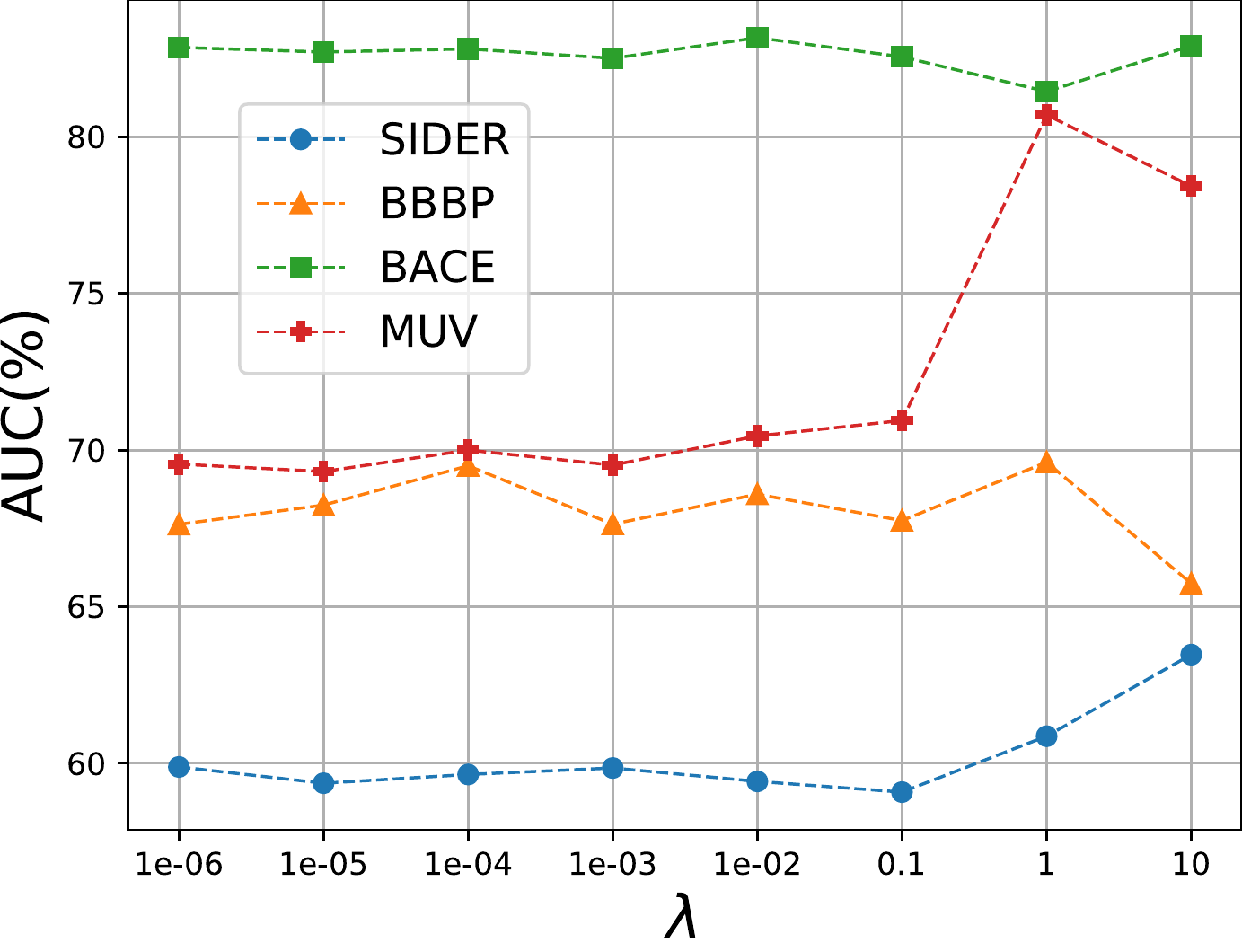}
    \end{minipage}%
    }%

\centering

\caption{Effect of hyper-parameter $\lambda$.}
\label{fig:sensitivity_lambda_GTOT}
\end{figure*}

Figure \ref{fig:sensitivity_lambda_GTOT} shows test AUC w.r.t. different $\lambda$'s on SIDER, BBBP, BACE and MUV, where the backbone is GIN~(contextpred). The performance benefits from a proper selection of $\lambda$ (from 0.001 to 10 in our experiments). When $\lambda$ is too small, the GTOT regularization term does not work; if it is too large, the Cross entropy term would be neglected(appears on BBBP and MUV), leading to performance degradation.

\subsection{Runtimes}\label{subsec:runtimes}
 
 We report 

the average training epoch time of different fine-tuning methods in Table \ref{tab:runtimes}. We note that our method is between 1 to 3 times slower than the vanilla Fine-Tuning baseline
This is caused by the frequent calls to the MOT solver. On the other hand, our method has a competitive speed-up compared with other methods. %
\begin{table*}[th]
\def\p{$\pm$} 
\setlength\tabcolsep{4pt} 
\centering
\vspace{-2mm}
\caption{The average fine-tuning training time per epoch with different methods is shown below and timings are measured in seconds. The float in parentheses is the standard deviation. )}
\vspace{-3mm}
\scalebox{0.84}{
\begin{tabular}{l|llllllll}
\toprule 
Methods & BBBP & Tox21 & Toxcast& SIDER & ClinTox & MUV & HIV & BACE \\
\midrule
Fine-Tuning~(baseline) & 5.22$\pm$0.47 & 13.95$\pm$0.65 & 11.34$\pm$1.19  & 3.26$\pm$0.20 & 2.65$\pm$0.19 & 43.39$\pm$0.31 & 30.49$\pm$8.52& 6.80$\pm$0.32 \\
\midrule
L2\_SP~\cite{xuhong2018explicit} & 5.30$\pm$0.89 & 23.79$\pm$1.36 & 27.34$\pm$1.05 & 5.32$\pm$0.47 & 6.76$\pm$0.72 & 227.32$\pm$29.41 & 136.47$\pm$7.44 & 4.26$\pm$1.07  \\

DELTA\cite{li2018delta} & 5.74$\pm$0.32 & 11.42$\pm$1.25  & 11.06$\pm$0.25 & 5.77$\pm$0.63 & 3.18$\pm$0.20 & 113.62$\pm$11.92 & 123.79$\pm$7.89 & 5.03$\pm$1.19 \\

Feature(DELTA w/o ATT) & 6.16$\pm$1.54 & 20.84$\pm$2.72 & 14.40$\pm$0.68 & 6.84$\pm$0.29 & 2.65$\pm$0.56 & 293.46$\pm$4.90 & 39.38$\pm$2.05 & 76.3$\pm$1.1  \\

BSS~\cite{chen2019catastrophic} & 6.20$\pm$0.63 & 30.84$\pm$0.54 & 6.52$\pm$0.57 & 24.92$\pm$2.03 & {70.9$\pm$5.1} & 316.89$\pm$3.38 & 126.59$\pm$3.32 & 6.05$\pm$1.70   \\

StochNorm~\cite{kou2020stochastic} &4.54$\pm$0.45 & 17.84$\pm$0.28 & 31.11$\pm$2.38 & 3.29$\pm$0.51 & 3.29$\pm$0.37 & 251.43$\pm$86.79 & 79.59$\pm$11.15 & 3.82$\pm$0.25 \\
\midrule
GTOT-Tuning~(Ours) & 5.68$\pm$0.90  & 22.74$\pm$0.75 & 34.12$\pm$6.49 & 4.64$\pm$0.30 & 2.95$\pm$0.30 & 71.45$\pm$5.83 & 73.37$\pm$29.97 & 11.26$\pm$0.57  \\
\bottomrule
    \end{tabular}
    }
    \vspace{-2mm}
    \label{tab:runtimes}    
\end{table*}


\begin{figure*}[htbp]
\centering
    \subfigure{
    \begin{minipage}[s]{0.7\linewidth}
    \centering
    \includegraphics[width=1\linewidth]{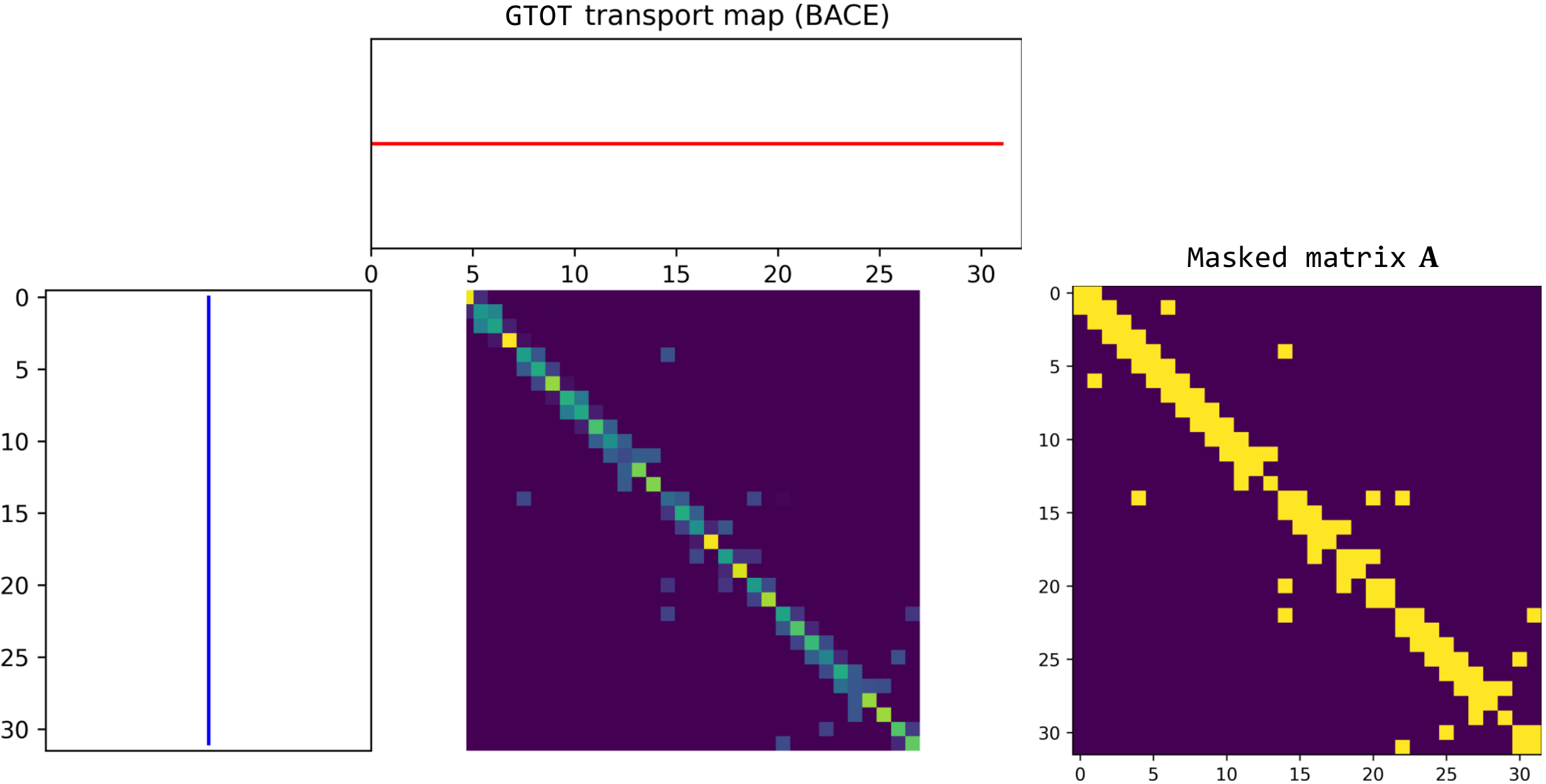}
    \caption*{(a)}
    \end{minipage}%
    }%
    
    \subfigure{
    \begin{minipage}[s]{0.7\linewidth}
    \centering
    \includegraphics[width=1\linewidth]{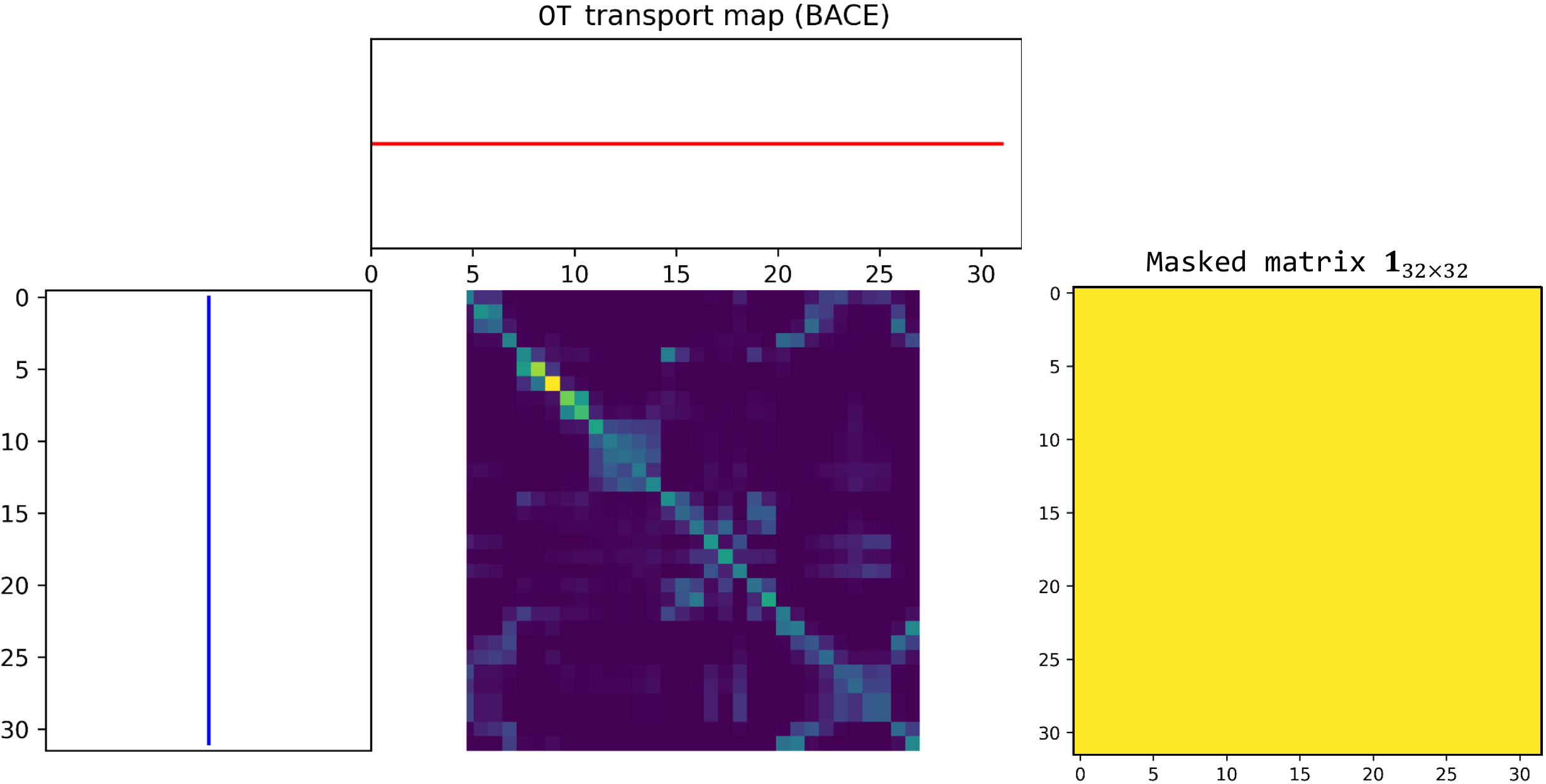}
    \caption*{(b)}
    \end{minipage}%
    }%
\centering

\caption{Comparison of GTOT~(MWD($\A$)) and OT 's (MWD($\mbf{1}_{32\times 32}$)) transport maps.Although when solved exactly, typical OT yields a sparse solution $\P^*$ containing $(2\max(n, m)-1)$ non-zero elements at most~[De Goes et al., 2011], the actual solution solved by iterative algorithm may tend to be dense. When the mask matrix is the adjacency matrix of a given graph, the transport plan is restricted to neighboring nodes and a more sparse transport map is obtained. Another important observation is that the values of the main diagonal of both are large, and GTOT retains most of the important values, removing unnecessary transfer processes. }
\label{fig:mwd_transport_map}
\end{figure*}

\begin{figure*}[htbp]
\centering
    \subfigure{
    \begin{minipage}[s]{0.7\linewidth}
    \centering
    \includegraphics[width=1\linewidth]{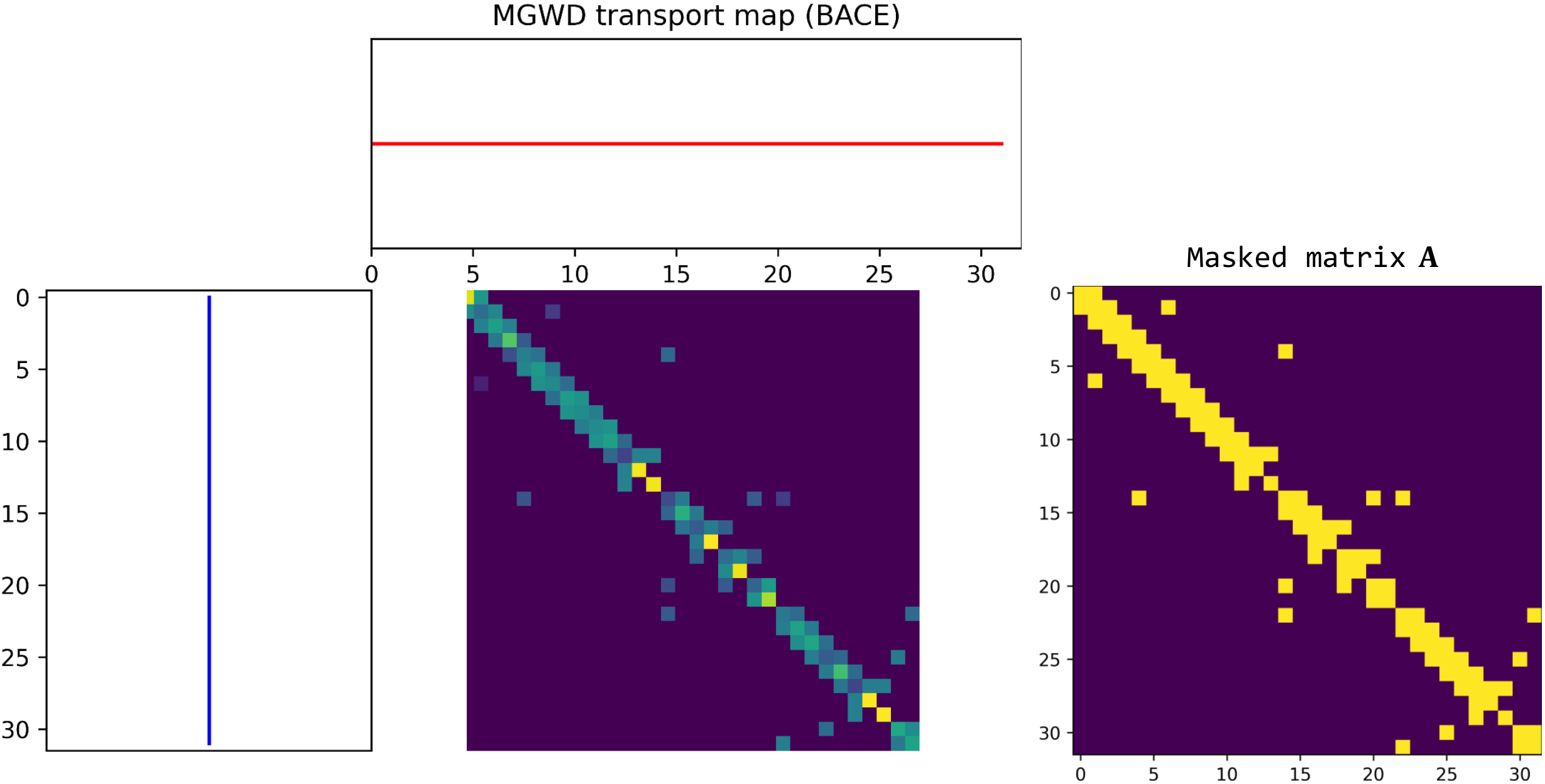}
    \caption*{(a)}
    \end{minipage}%
    }%
    
    \subfigure{
    \begin{minipage}[s]{0.7\linewidth}
    \centering
    \includegraphics[width=1\linewidth]{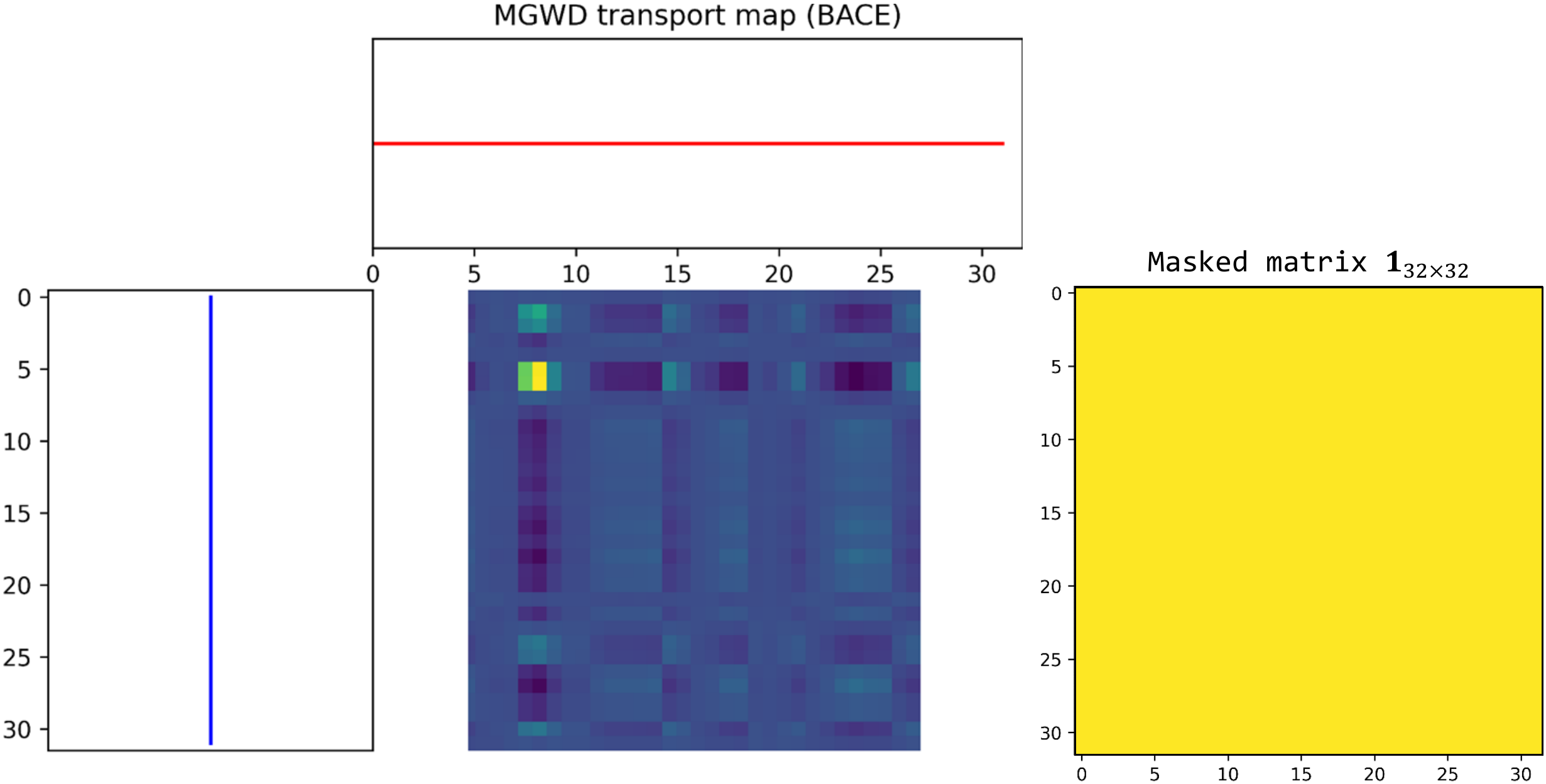}
    \caption*{(b)}
    \end{minipage}%
    }%

\centering

\caption{Comparison of MGWD~($\M=\A$) and GWD's($\M=\mbf{1}_{32\times 32}$) transport maps. When masked matrix is the adjacency matrix of a given graph, the transport plan is restricted to neighboring nodes and a more sparse transport map is obtained.   }
\label{fig:mgwd_transport_map}
\end{figure*}

\end{document}